\documentclass{article}

\newif\ifanonymous
\anonymousfalse
\ifanonymous
\usepackage[dblblindworkshop]{neurips_2026}
\else
\usepackage[dblblindworkshop, final]{neurips_2026}
\fi

\usepackage[utf8]{inputenc} 
\usepackage[T1]{fontenc}    
\usepackage{hyperref}       
\hypersetup{pdftitle={Harness as a Language: A Minimalist Agent Framework With Maximal Expressivity}}
\ifanonymous\else
\hypersetup{pdfauthor={Zhening Li, Joshua Liu, Mateja Vukelic, Nicole Shen, Supriya Lall, Amitayush Thakur, Alex Zhang, Omar Khattab, Jonathan Light, Armando Solar-Lezama}}
\fi
\usepackage{url}            
\usepackage{booktabs}       
\usepackage{amsfonts}       
\usepackage{nicefrac}       
\usepackage{microtype}      
\usepackage{xcolor}         
\usepackage{wrapfig}

\ifdefined\pdfsuppressptexinfo\pdfsuppressptexinfo=-1\fi

\usepackage{graphicx}  
\usepackage{multirow}
\usepackage{listings}
\usepackage{amsmath}
\usepackage{mathtools}
\usepackage{tikz}  
\usepackage{multicol}  
\usepackage{cleveref}  
\usepackage{subcaption}  
\usepackage{bussproofs}  
\usepackage{enumitem}  
\usepackage{proof}
\usepackage{listings}
\usepackage{makecell}
\usepackage{graphicx}   
\usepackage{tabularx}
\usepackage{wrapfig}    
\usepackage{tcolorbox}
\tcbuselibrary{breakable}
\usepackage{amsthm}
\usepackage{pifont}

\newcommand{\xmark}{\ding{55}}%

\definecolor{low}{RGB}{203, 72, 120} 
\definecolor{medium}{RGB}{33, 144, 141} 
\definecolor{high}{RGB}{94, 201, 98} 

\newcommand{\oursimpl}{JAZ}

\definecolor{jaz-paper}{HTML}{FBF8F6}
\definecolor{jaz-paper-alt}{HTML}{F4EFEB}
\definecolor{jaz-berry}{HTML}{B6397A}
\definecolor{jaz-accent}{HTML}{105C66}

\lstdefinestyle{bare}{
  language={},
  emph={},
  basicstyle=\ttfamily\tiny,
  keywordstyle={},
  commentstyle={},
  stringstyle={},
  identifierstyle={},
  showstringspaces=false,
}

\newlist{titledlist}{description}{1}
\setlist[titledlist]{
  style=nextline,   
  labelsep=0pt,     
  leftmargin=0pt,   
  labelindent=0pt,
  itemindent=2em,   
  listparindent=2em,
  font=\bfseries,   
  parsep=0pt,
  itemsep=.7\baselineskip
}

  {\list{}{\leftmargin=0.25in\rightmargin=0.0in}\item[]}%
  {\endlist}
  {\list{}{\leftmargin=0.5in\rightmargin=0.0in}\item[]}%
  {\endlist}

\newif\ifcomments
\commentstrue

\ifcomments
  \newcommand{\jl}[1]{\textcolor{red}{[JL: #1]}}
\else
  \newcommand{\jl}[1]{}
\fi

\ifcomments
  \newcommand{\zl}[1]{\textcolor{blue}{[ZL: #1]}}
\else
  \newcommand{\zl}[1]{}
\fi

\tcbuselibrary{breakable}
\newtcolorbox{promptbox}[1][]{
  breakable,
  colback=jaz-paper-alt,
  colframe=jaz-paper-alt,
  boxrule=0pt,
  arc=3mm,
  left=3mm,
  right=3mm,
  top=2mm,
  bottom=2mm,
  #1
}

\newenvironment{items}
  {\begin{itemize}[leftmargin=*,itemsep=2pt,topsep=0pt,parsep=0.4pt,partopsep=0pt]}
  {\end{itemize}}

\newenvironment{enum}
  {\begin{enumerate}[leftmargin=*,itemsep=2pt,topsep=0pt,parsep=0.4pt,partopsep=0pt]}
  {\end{enumerate}}

\newif\ifpreprint
\preprinttrue
\newif\ifmetaagents
\metaagentstrue

\newif\ifttcl
\ttclfalse
\newif\ifiab
\iabfalse
\ifmetaagents\ttclfalse\iabfalse\fi

\ifttcl
\workshoptitle{Towards Test-Time Continual Learning Agents (TTCL)}
\fi
\ifiab
\workshoptitle{Interpreting Agent Behavior (IAB)}
\fi

\ifmetaagents
\workshoptitle{Managing Agents that Manage Agents:
  Workshop on Responsible Use of Meta-Agents}
\makeatletter
\renewcommand{\@trackname}{\@workshoptitle\ (NeurIPS \@neuripsyear)}
\makeatother
\fi

\title{Harness as a Language: A Minimalist Agent Framework With Maximal Expressivity}

\author{%
  Zhening Li$^1$ \quad Joshua Liu\thanks{Equal contribution}$^{*1}$ \quad Mateja Vukelic$^{*1}$ \\
  \textbf{Nicole Shen$^1$ \quad Supriya Lall$^1$ \quad Amitayush Thakur$^2$} \\
  \textbf{Alex Zhang$^1$ \quad Omar Khattab$^1$ \quad Jonathan Light$^2$ \quad Armando Solar-Lezama$^1$} \\
  $^{1}$MIT CSAIL \quad $^{2}$Independent Researcher
}

\begin{document}

\maketitle

\begin{center}
    \centering
    \vspace{-0.35in}
    \includegraphics[width=0.8\linewidth]{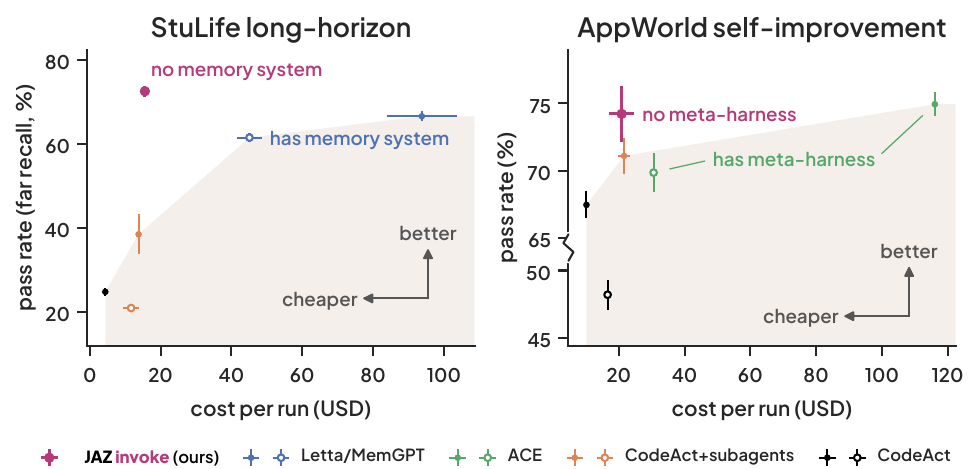}
    \label{fig:combined_cost}
\end{center}

\begin{abstract}
  Modern language-model agents are built around the \textit{agent loop}: the LLM
is placed in an environment exposing a set of tools, and the LLM has full control over
the workflow by alternating between tool calls and observing their output.
However, certain capabilities such as long-term memory and self-improvement
currently require specialized systems beyond the agent loop itself.
We built an LLM agent framework, \oursimpl{},
to explore the extent to which a minimal harness that is little more than the
\textit{agent loop itself} can accomplish tasks these specialized systems are built for.
\oursimpl{} exposes a single LLM-based primitive
\lstinline|invoke| and provides a set of built-in hooks that allow the programmer
to apply constraints and perform monitoring. Generalizing existing code-mode agent loops, \lstinline|invoke| is
the simplest loop that satisfies two defining properties:
(1) the LLM can write arbitrary executable code that can include recursive calls to \lstinline|invoke|;
(2) all inputs to \lstinline|invoke| and the LLM's interaction history with the code environment
are variables in the code environment, i.e., everything visible to the LLM can be accessed programmatically.
We motivate our design from first principles, viewing \lstinline|invoke|
as a language primitive representing a function whose implementation is provided
at runtime by an LLM every time it is called.
To validate the design of our core \lstinline|invoke| primitive,
we evaluate \lstinline|invoke| --- with only prompting, no manually designed
tools, harness, or external systems (e.g., memory or the file system) --- on workflows traditionally implemented
through specialized harnesses.
On long-horizon workflows requiring recall far beyond the context window,
\oursimpl{} \lstinline|invoke| outperforms Letta/MemGPT
at a fraction of its cost on the recall-heavy portion of StuLife.
On continual self-improvement,
\oursimpl{} \lstinline|invoke| matches ACE at a fraction of its cost
on AppWorld.
\ifanonymous
Our code is located at 
\url{https://anonymous.4open.science/r/jaz-evals-iclr-6151}.
\else
The \oursimpl{} framework source code is located at \url{https://github.com/jaz-lang/jaz}
and our evaluation code is located at \url{https://github.com/jaz-lang/jaz-evals}.
\fi

\end{abstract}

\ifpreprint
\begin{wrapfigure}[25]{r}{0.48\linewidth}
    \centering
    \vspace{-0.55in}
    \includegraphics[width=1.0\linewidth]{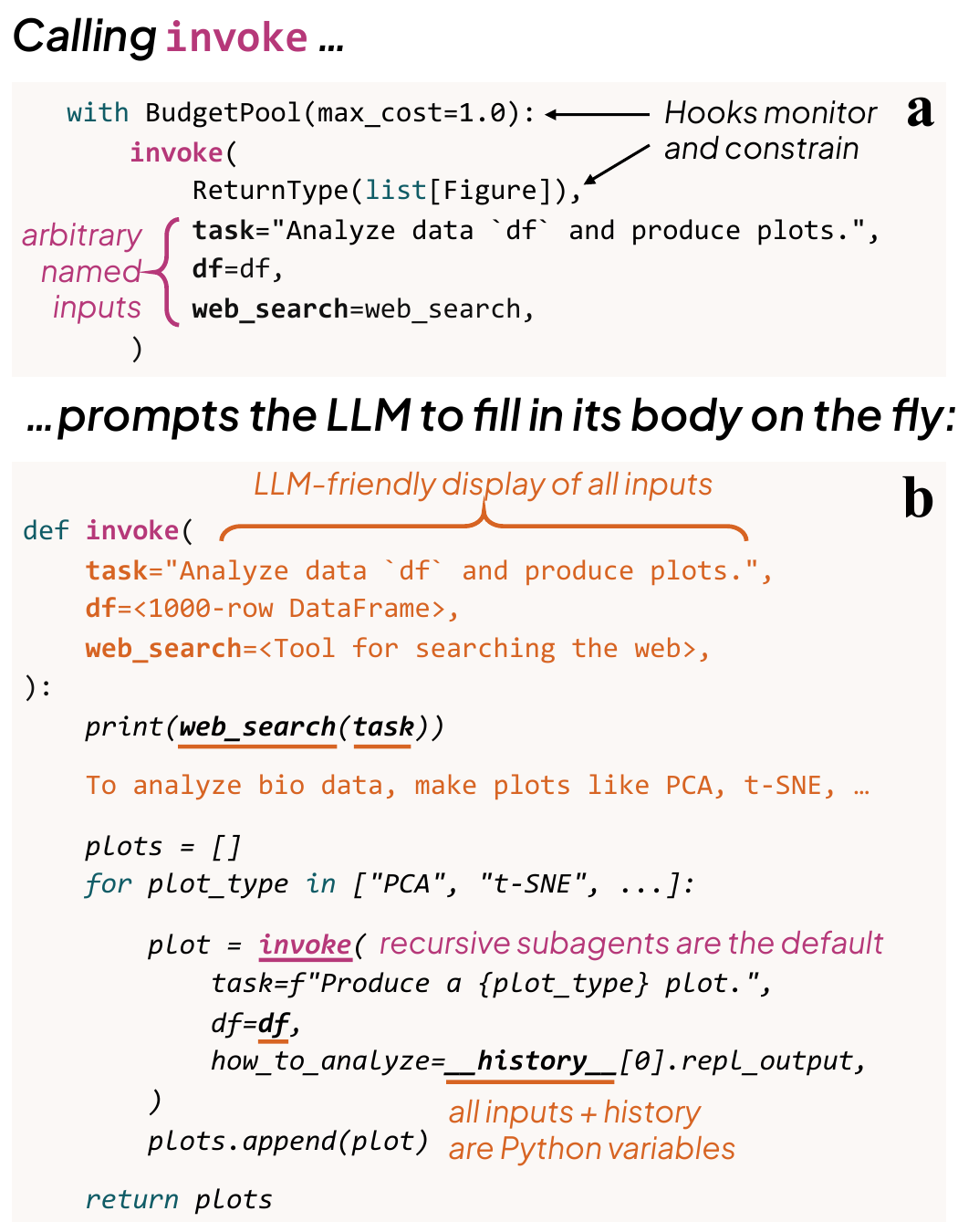}
    \vspace{-0.2in}
    \caption{\textbf{In \oursimpl{}, \lstinline|invoke| acts as a function whose body is written by an LLM in a REPL every time it's called. All named inputs to \lstinline|invoke| and the REPL history itself are variables available in the REPL.}}
    \label{fig:teaser}
\end{wrapfigure}
\else
\begin{wrapfigure}[19]{r}{0.45\linewidth}
    \centering
    \vspace{-0.6in}
    \includegraphics[width=1.0\linewidth]{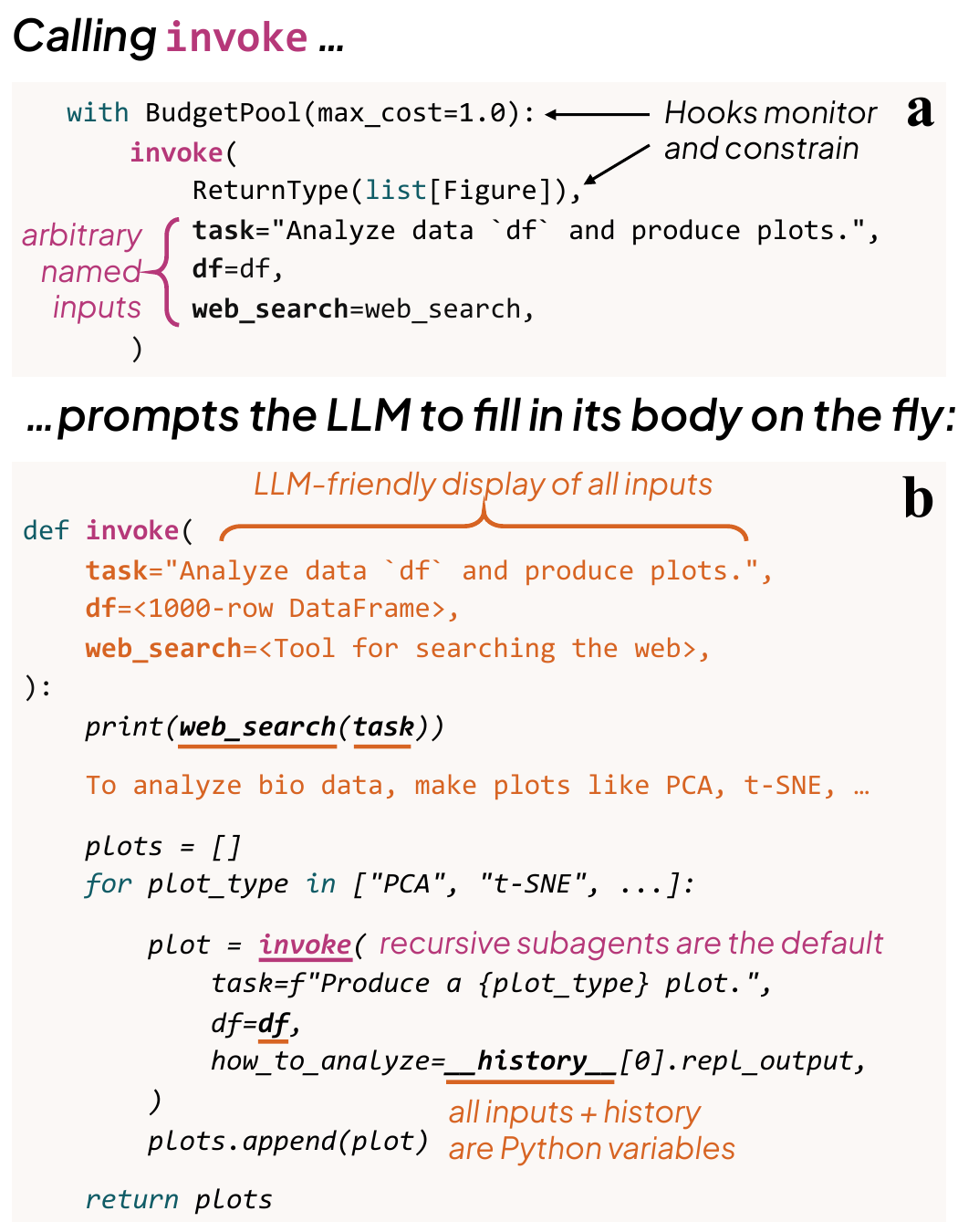}
    \vspace{-0.2in}
    \caption{\oursimpl{} \lstinline|invoke|.}
    \label{fig:teaser}
\end{wrapfigure}
\fi

\section{Introduction}

\vspace{-0.05in}

Language model agents are typically built around a core \textit{agent loop} in which the model repeatedly interacts with an environment through actions and observations.
Increasingly expressive agent loop paradigms have been proposed over time.
Earlier ReAct~\citep{yao2022react} gave the model full freedom to decide
which action to take at every step of a workflow.
CodeAct~\citep{wang2024codeact} generalized it by providing
the model the expressivity to orchestrate workflows with executable code in a REPL,
and typical implementations \citep{chase2023langchain,roucher2025smolagents}
also allow the user to provide Python objects the agent can interact with in its REPL.
Increasing expressivity has unlocked new capabilities.
Most notably, the RLM~\citep{zhang2025rlm} demonstrated that CodeAct+subagents,
without any external tools or harness, could outperform LLMs on long-context tasks by
providing the input to the agent as a Python object it can manipulate in its REPL.
However, agentic capabilities such as long-horizon memory and self-improvement
remain implemented through specialized external harnesses layered on top of these loops.

We study whether a minimal harness that is little more than the
\textit{agent loop itself} can exhibit these capabilities.
For this, we built an LLM agent framework, \oursimpl{},
whose core primitive \lstinline|invoke|
is a minimal language-level abstraction that generalizes CodeAct \citep{wang2024codeact} and RLMs \citep{zhang2025rlm} (\Cref{fig:teaser}).
First, the model writes arbitrary executable code in this \lstinline|invoke|-augmented language,
allowing tools and recursive subagents to be orchestrated through ordinary program control flow.
In other words, like RLMs, recursive subagents are the default.
Second, \textit{everything} visible to the model is also a variable in the code environment.
While modern implementations of CodeAct and RLMs
let the user pass data and tools into the agent's REPL,
\lstinline|invoke| additionally makes the user prompt and REPL history variables in the REPL.

On top of the \lstinline|invoke| primitive, \oursimpl{} provides utilities that allow a user
to monitor and constrain agent behavior. Observability hooks log agent trajectories,
validation hooks validate return conditions, and resource constraint hooks impose budgets
on cost, iterations, and context window usage.
Although hooks are not essential to the core \oursimpl{} primitive,
they allow users of the \oursimpl{} framework to observe and exercise control over their agent.

We show that \oursimpl{}, with little more than pure prompting,
can realize agentic workflows traditionally implemented through specialized harnesses.
On long-horizon tasks requiring recall far beyond the model's context window,
\oursimpl{} \lstinline|invoke| is prompted to delegate to a subagent whenever its context runs out,
while keeping a reference to the full conversation history.
With GPT-5.4 nano, it achieves 73\% on the recall-heavy subset of StuLife~\citep{cai2026stulife},
outperforming both CodeAct+subagents (39\%)
and the specialized long-horizon harness Letta/MemGPT \citep{packer2023memgpt} (62--67\%),
at 0.2--0.3x the latter's cost.
On AppWorld \citep{trivedi2024appworld},
\oursimpl{} \lstinline|invoke| self-improves across the sequence of
test tasks by iteratively updating its prompt and tools inside its REPL based on test feedback.
With a GPT-5.4 top-level \lstinline|invoke| and GPT-5.4 nano sub-invokes,
it achieves 74\%, outperforming CodeAct (68\%) and CodeAct+subagents (71\%),
and matching a specialized self-improvement harness, ACE \citep{zhang2025ace} (70--75\%),
at 0.2--0.7x its cost.
Together, these results show that prompting a minimal agent loop is sufficient to
elicit long-horizon and self-improvement capabilities,
rather than requiring specialized external harness components.


Our main contributions are:
\begin{itemize}[leftmargin=*,itemsep=2pt,topsep=0pt,parsep=0.4pt,partopsep=0pt]
    \item We introduce and formalize \lstinline|invoke|,
    an LLM-based programming language construct that represents a function
    whose implementation is provided by the LLM at call time.

    \item We introduce JAZ, an agent framework whose core primitive is the \lstinline|invoke| agent loop.
    An expressive hook system enables utilities such as observability, budget control, validation,
    and general modifications to the agent loop.

    \item We show empirically that the additional expressivity of
    \oursimpl{} \lstinline|invoke| enables
    long-horizon and self-improvement workflows without external systems, tools, or harness components.
    With prompting, \oursimpl{} \lstinline|invoke| outperforms CodeAct+subagents
    at a similar cost, and matches or outperforms specialized harnesses at a
    substantially lower cost.
\end{itemize}

\ifttcl
Related work is described in \Cref{sec:related}.
\else\ifiab
Related work is described in \Cref{sec:related}.
\fi
\fi

\section{\texorpdfstring{\lstinline[basicstyle=\ttfamily\large]|invoke|}{invoke}: An LLM-based language primitive}


To motivate the design of \oursimpl{},
we consider a simple and general way to extend any existing programming language with an LLM-based primitive.
We call this primitive \textit{\lstinline|invoke|}.
\ifttcl
Discussion on its minimality and generality is given in \Cref{sec:minimal}.
\else\ifiab
Discussion on its minimality and generality is given in \Cref{sec:minimal}.
\fi
\fi

\subsection{Functional \texorpdfstring{\lstinline|invoke|}{invoke}}

\begin{figure}[tb]
    \centering
    \vspace{-0.2in}
    \includegraphics[width=\linewidth]{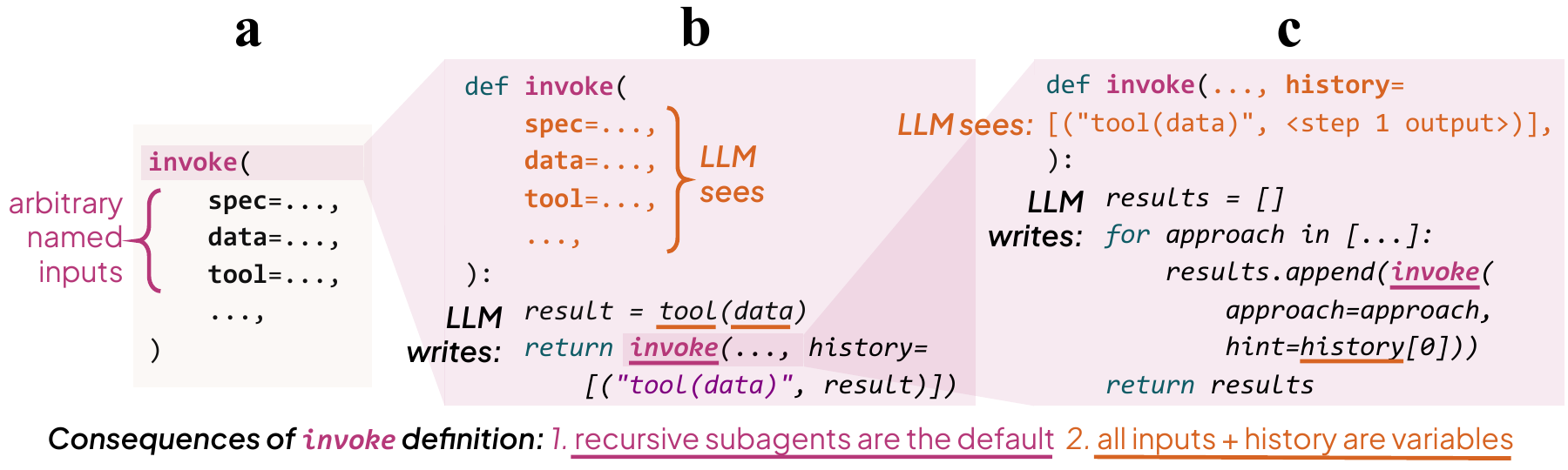}
    \vspace{-0.2in}
    \caption{In the functional version of \lstinline|invoke|, the LLM writes a complete implementation
    for every invocation. A closed agent loop arises from tail-recursive \lstinline|invoke|.}
    \vspace{-0.2in}
    \label{fig:functional_invoke}
\end{figure}

\Cref{fig:functional_invoke} shows an example execution of the \textit{functional} version of
the \lstinline|invoke| primitive.
\lstinline|invoke| is thought of as a function that can take as inputs
arbitrary named arguments. These inputs can serve various purposes.
Some are inputs in the traditional sense (\lstinline|df| in \Cref{fig:teaser}a,
\lstinline|data| in \Cref{fig:functional_invoke}a),
some are the specification (\lstinline|task| in \Cref{fig:teaser}a,
\lstinline|spec| in \Cref{fig:functional_invoke}a),
and some are tools the agent can call (\lstinline|web_search| in \Cref{fig:teaser}a,
\lstinline|tool| in \Cref{fig:functional_invoke}a).
However, \lstinline|invoke| treats all inputs the same and makes
no distinction between prompts, tools, and ``regular'' inputs.

Unlike a regular function, whose implementation is provided
in the static code, \lstinline|invoke| has its implementation
generated by the model every time it is called (\Cref{fig:functional_invoke}b).
The model sees these inputs as a compact string representation,
and writes code that fills in the body of \lstinline|invoke|.
Note that this code is written in the same programming language,
which has been augmented with the \lstinline|invoke| primitive.
\ifttcl
\Cref{sec:minimal} elaborates on the minimality and generality of the primitive.
\fi
For a formal treatment in lambda calculus, see \Cref{sec:formal}.


\subsection{The \texorpdfstring{\lstinline|invoke|}{invoke} agent loop}

In principle, functional \lstinline|invoke| is already at least as expressive as a closed agent loop,
which can be expressed as tail-recursive \lstinline|invoke| (\Cref{fig:functional_invoke}b--c).
To emulate a closed agent loop where the model sees the output
of its code and decides its next action based on that,
it could write a tail-recursive call to \lstinline|invoke|
that passes in the inputs, any local variables, and the interaction history with the agent loop (\Cref{fig:functional_invoke}b).
This delegates to a subagent whose context
contains the same inputs but also the history so far (\Cref{fig:functional_invoke}c),
thus behaving as the next step of the agent loop.

In practice, however, this approach is inefficient: it wastes memory in Python,
and benefiting from LLM caching optimizations requires an exact prefix relationship among
consecutive LLM queries.
Thus, instead of having the model write the tail-recursive call,
we can let the harness automatically append it and apply tail-call optimization, resulting in
an actual code-mode agent loop where the LLM writes code in a REPL.
This agent loop has two distinguishing properties that are
direct consequences of the definition of functional \lstinline|invoke| (\Cref{fig:teaser,fig:functional_invoke}):

\begin{enumerate}[leftmargin=*,itemsep=2pt,topsep=0pt,parsep=0.4pt,partopsep=0pt]
    \item The \textit{language} of the model's output is the augmented language,
    which already contains \lstinline|invoke|.
    Thus, by default, the model can write not just arbitrary
    code in the original language, but also code that calls \lstinline|invoke|,
    which acts as subagents. We call these recursive \lstinline|invoke| calls
    \textit{sub-invokes}.
    
    \item \textit{Everything} the language model sees is a variable that it
    can reference in its code.

    \begin{itemize}[leftmargin=*,itemsep=2pt,topsep=0pt,parsep=0.4pt,partopsep=0pt]
        \item This includes \textit{all} inputs to \lstinline|invoke|, treating all inputs equally
        regardless of their purpose. Thus, an input intended as the prompt
        can be accessed as a variable, just as an input intended as a tool.
        \item The \textit{REPL history itself} is also a variable in the REPL.
    \end{itemize}
\end{enumerate}

Property 1 makes \lstinline|invoke| a variant of CodeAct+subagents or RLMs
\citep{wang2024codeact,zhang2025rlm}.
Property 2 --- especially the ``REPL history is a REPL variable''
aspect --- distinguishes \lstinline|invoke| from existing CodeAct-like agents
(e.g., smolagents \citep{roucher2025smolagents}, RLMs \citep{zhang2025rlm}),
which treat the user prompt and REPL history only as a list of messages shown to the agent,
not as variables the agent can programmatically interact with.

We call an agent loop satisfying the two properties above an
\textit{\lstinline|invoke| agent loop}.

\ifttcl\else\ifiab\else
\ifttcl
\section{\texorpdfstring{\lstinline[basicstyle=\ttfamily\large]|invoke|}{invoke} is minimal and general}
\else\ifiab
\section{\texorpdfstring{\lstinline[basicstyle=\ttfamily\large]|invoke|}{invoke} is minimal and general}
\else
\subsection{\texorpdfstring{\lstinline|invoke|}{invoke} is minimal and general}
\fi
\fi
\label{sec:minimal}

Here, we illustrate informally how functional \lstinline|invoke| arises naturally from considering the \textit{minimal} LLM primitive that
can be generally introduced to \textit{any} Turing-complete language.
The usual LLM primitive takes a string input and outputs a string, and that is arguably
the minimal LLM primitive in programming languages that \textit{have strings}.
However, the simplest languages, such as lambda calculus, do not have a string data type and thus
do not admit the usual LLM primitive. Nonetheless, \textit{the terms of a language themselves are strings},
irrespective of the language.
So the LLM can simply take terms in the language as inputs, and output
a term in the language. If the output term can reference inputs as variables, then we arrive
exactly at the definition of functional \lstinline|invoke|. Its syntax is that of a
function call that takes arbitrary named inputs, and the function call's semantics
is to query an LLM to generate the body of the function, which is executed.
Thus, the definition of \lstinline|invoke| arises naturally from modifying the usual LLM primitive
to be generally applicable to all Turing-complete languages.
Furthermore, as shown in \Cref{sec:formal}, extending a language with \lstinline|invoke|
introduces just one new piece of syntax and one additional rule to the formal semantics,
so \lstinline|invoke|'s formal definition exemplifies its minimality as well.

\fi\fi

\section{\oursimpl{}: An LLM agent framework based on \texorpdfstring{\lstinline[basicstyle=\ttfamily\large]|invoke|}{invoke}}  \label{sec:jaz}

We built \oursimpl{}, an agent framework that implements the
\lstinline|invoke| agent loop as its core language model primitive.
\oursimpl{} additionally implements the following systems to make
it easier for users to build agents using the \lstinline|invoke| primitive.

\paragraph{Input serialization.} Naively, each input to \lstinline|invoke| is displayed
to the LLM as its \lstinline|str()| serialization. However, Python's default \lstinline|str()|
has two major issues for this purpose: (a) its size is unbounded, so large objects would blow
the model's context window; and (b) it contains no useful information for functions, methods,
and objects that don't define their own custom string representation.
\oursimpl{} addresses (a)
by truncating to a configurable length (50,000 characters by default). For (b), we implement
our own serialization function. In the simplified picture, a function shows its signature and
docstring; a class or module shows its docstring, public attributes, and methods;
for any other object, we use \lstinline|str()|
if it defines its own custom string representation,
else we show the object's public attributes and its class's docstring and methods.
Details can be found in our source code.

\paragraph{Dynamic scoping.}
Instead of requiring that every input to \lstinline|invoke| be provided as an explicit
argument, we provide a primitive \lstinline|scope| that creates a
scope of variables where \lstinline|invoke| calls within the scope automatically receive
those variables, including all their recursive sub-invokes. In other words,
variables specified by \lstinline|scope| are subject to \textit{dynamic scoping} \citep{hanson2001dynamicscoping}
for \lstinline|invoke| calls.
In \oursimpl{}, \lstinline|scope| takes the form of a Python context manager.
The most common use case is to make a tool available not just to the top-level agent, but also
to all its recursive subagents:
\begin{lstlisting}[basicstyle=\ttfamily\small, upquote=True]
from jaz import invoke, scope

def web_search(...):
    ...

with scope(web_search=web_search):
    # Top-level agent and all subagents can use `web_search`
    invoke(task="Use subagents to find 100 agent framework papers.")
\end{lstlisting}

Although traditional programming prefers static scoping over dynamic scoping \citep{steele1978artofinterpreter},
only dynamic scoping is appropriate for \lstinline|invoke|.
Static scoping enables reusing a function in different contexts without surprising changes
in behavior due to a changing context. However,
\lstinline|invoke| does not have a static scope as it does not have a static implementation,
and there is no reuse as every call to \lstinline|invoke| generates a new implementation.

\paragraph{Hooks.} Hooks allow a user to extend \lstinline|invoke| by providing
callbacks that observe or influence the agent as it is running. \oursimpl{}
provides the following built-in hooks:
\begin{itemize}[leftmargin=*,itemsep=2pt,topsep=0pt,parsep=0.4pt,partopsep=0pt]
    \item \textit{\textbf{Observability:}} \lstinline|PrintLogger|, \lstinline|FileLogger|,
    \lstinline|TrajectoryDirectoryRecorder|, \lstinline|TrajectoryRecorder|,
    \lstinline|LangfuseTracing|, \lstinline|JaegerTracing|.
    \item \textit{\textbf{Resumability:}} \lstinline|TrajectoryRecorder|, \lstinline|TrajectoryReplay|.
    \item \textit{\textbf{Resource control:}} \lstinline|BudgetPool|, \lstinline|IterationLimit|,
    \lstinline|RecursionLimit|, \lstinline|BudgetForcing|, \lstinline|ContextWindowWarning|.
    \item \textit{\textbf{Validation:}} \lstinline|ReturnType|, \lstinline|ValidateReturn|,
    \lstinline|ValidateREPLCode|.
\end{itemize}
Users can also extend \oursimpl{} by writing their own custom hooks (\Cref{app:extensibility}).

Just like input variables, hooks can be either local to an \lstinline|invoke|
or be subject to dynamic scoping. Hooks passed as explicit positional
arguments to \lstinline|invoke| are local, while a hook activated
as a context manager is scoped. In the following example,
the scoped \lstinline|BudgetPool(max_cost=5)| hook applies a budget of \$5
shared across all \lstinline|invoke| calls under the context manager,
including both top-level calls and all their recursive sub-invokes.
On the other hand, the local \lstinline|ReturnType(float)| enforces
the return type \lstinline|float| only for its top-level \lstinline|invoke| and not
for its sub-invokes.
\begin{lstlisting}[basicstyle=\ttfamily\small]
from jaz import invoke
from jaz.hooks import ReturnType, BudgetPool

with BudgetPool(max_cost=5):  # scoped hook
    invoke(
        ReturnType(float),  # local hook
        question="What is the value of ...?"
    )
    invoke(another_question="Why is ...?")
\end{lstlisting}

\section{Experiments} \label{sec:experiments}

In this section, we explore how the expressivity of \oursimpl{} \lstinline|invoke|
allows us to replace specialized
external harnesses with prompting for two types of workflows: (1)
long-horizon workflows requiring recall beyond the context window;
and (2) continual learning and self-improvement across a sequence of tasks.
These workflows are traditionally targeted with external harness components,
such as memory systems for long-range recall, and specialized self-improving systems
for continual self-improvement. Here, we use \oursimpl{} to demonstrate that a minimal LLM-based primitive,
the \lstinline|invoke| agent loop, can perform these workflows as well
without external systems, tools, or harness components.

We compare \lstinline|invoke| with agent loop paradigms under a similar prompt-only setup
(CodeAct and CodeAct+subagents \citep{wang2024codeact,zhang2025rlm}), described below.
We also compare with domain-specific harnesses (Letta/MemGPT \citep{packer2023memgpt}
and ACE \citep{zhang2025ace}), described in the individual subsections
(\Cref{sec:long_horizon,sec:csi}).
\begin{items}
    \item \textbf{CodeAct} \citep{wang2024codeact} is an agent loop paradigm
    generalizing ReAct \citep{yao2022react} tool calling where the model's action
    is not a single tool call, but a piece of code that orchestrates tool calls.
    Implementations \citep{chase2023langchain,roucher2025smolagents}
    involve a code REPL that is seeded with user-provided tools and objects,
    and the LLM writes code in the REPL in a closed loop.
    The core difference from the \lstinline|invoke| agent loop is that the
    user prompt and the REPL history are only shown to the LLM and not
    available as variables in the REPL.
    \item \textbf{CodeAct+subagents} is the agent loop paradigm
    that augments CodeAct with recursive subagents,
    where ``subagent'' is understood to mean generalist subagents
    instead of specialized subagents with designated roles.
    This paradigm was popularized by the RLM work \citep{zhang2025rlm},
    which implemented a version where the REPL is seeded with a large string,
    making the agent act as a long-context language model.
\end{items}

For a controlled comparison,
we reimplement CodeAct and CodeAct+subagents in \oursimpl{} as controlled ablations of \lstinline|invoke|.
CodeAct+subagents is achieved by a custom hook that removes the user prompt and REPL history variables
from \lstinline|invoke|'s REPL and makes the minimal equivalent change to prompt templates.
CodeAct additionally uses the built-in \lstinline|RecursionLimit| hook to disable recursive subagents.
To validate our implementation, our long-horizon experiments also report results from
an established agent framework, smolagents \citep{roucher2025smolagents}.
However, because smolagents underperforms
our implementation due to confounds (defects in prompting and REPL implementation),
our comparisons between \oursimpl{} \lstinline|invoke| and CodeAct(+subagents)
will refer to the \oursimpl{} implementation of CodeAct(+subagents).
Both our environments involve completing a sequence of tasks, so a return guard
(\lstinline|ValidateReturn| in \oursimpl{}, \lstinline|final_answer_checks| in smolagents)
is used to make the agent keep working if it stops before all tasks have been attempted.
See \Cref{app:exp} for more details.

\subsection{Long-horizon with long-range recall} \label{sec:long_horizon}

Here, we focus on long-horizon environments with the following
properties:
\begin{itemize}[leftmargin=*,itemsep=2pt,topsep=0pt,parsep=0.4pt,partopsep=0pt]
    \item The task requires so many iterations (e.g., thousands) that the conversation history
    would exceed the LLM's context window multiple times if no
    subagents were used.
    \item Making the optimal decision frequently requires recalling
    some detail in the conversation history multiple context windows in the past.
    Compaction would lose the detail, and long-range dependencies make it difficult
    to cleanly decompose the workflow into subtasks.
\end{itemize}

In \oursimpl{}, when the LLM's context window fills up,
the agent can delegate the remainder
of the task to a subagent while losslessly passing in the entire conversation history
by reference:
\begin{lstlisting}[basicstyle=\ttfamily\small]
return invoke(
    ...,  # original inputs to the top-level invoke
    prev_history=__history__,  # REPL history
    prev_progress_summary=...,  # summary of history
    next_steps=...,  # next steps for the subagent
    ...,  # other state to keep track of
)
\end{lstlisting}
We call the above pattern \textit{tail-recursive delegation}.
In our experiments, we elicit this behavior with \oursimpl{}'s 
\lstinline|ContextWindowWarning| hook, which appends a user message
to the agent's conversation when its size passes a certain threshold.
The user message tells the agent to delegate all remaining work to a subagent
and shows a code template to follow (\Cref{app:long,app:context_window_warning}).

The general pattern above subsumes various forms of
long-horizon context management.
For example, \textit{compaction} \citep{wang2025compact,anthropic_compaction}
corresponds to passing in a \lstinline|prev_progress_summary| only.
Automated context management (e.g., Chroma Context-1~\citep{bashir2026context1})
corresponds to passing in a filtered version of
\lstinline|__history__| with certain entries removed.
More generally, access to \lstinline|invoke| gives the agent control
over the next LLM call's inputs and context,
whereas access to \lstinline|__history__|
and a code environment makes it efficient to build this context off of the agent's current context.

\ifpreprint
\begin{table}[tb]
\small
\begin{center}
\caption{Results on the long-horizon environment StuLife \citep{cai2026stulife}.
Methods marked ``per-task'' solve each StuLife task
individually with no state or memory persistence across tasks.
We report both pass rate (fraction of scored tasks with a perfect score) and average score,
for both the full scored set of 939 tasks (``all'') and the subset of 207 tasks that
require recall of information delivered over 50 tasks ago (``far recall'').
We use GPT-5.4 nano (high) across all methods.
Per-task methods report a 3-run mean and combined per-task SEM,
and all other methods report a 6-run mean and SEM to account for higher variance.
See \Cref{tab:long_horizon_full} for all individual data points.
}
\vspace{0.05in}
\label{tab:long_horizon}
\setlength{\tabcolsep}{5pt}
\begin{tabular}{rcccccc}
\toprule
 & & \multicolumn{2}{c}{StuLife (all)}
 & \multicolumn{2}{c}{StuLife (far recall)} \\
\cmidrule(lr){3-4} \cmidrule(lr){5-6}
& \smash{\scriptsize\shortstack{prompt-\\only?\,*}} & Pass (\%) & Score (\%) & Pass (\%) & Score (\%) & Cost (\$) \\
\midrule
CodeAct\textsubscript{\oursimpl{}} \ifpreprint\citep{wang2024codeact} \fi {\scriptsize (per-task)} & \checkmark & 52.5 $\pm$ 0.4 & 59.2 $\pm$ 0.3 & 24.8 $\pm$ 0.9 & 25.8 $\pm$ 0.9 & \textbf{4.4} $\pm$ 0.1 \\
CodeAct+subagents\textsubscript{smolagents \ifpreprint\citep{roucher2025smolagents}\fi} & \checkmark & 34.5 $\pm$ 4.8 & 37.8 $\pm$ 5.1 & 20.9 $\pm$ 0.4 & 22.0 $\pm$ 0.6 & \underline{11.7} $\pm$ 2.2 \\
CodeAct+subagents\textsubscript{\oursimpl{}} & \checkmark & 63.3 $\pm$ 2.0 & 71.8 $\pm$ 2.2 & 38.5 $\pm$ 4.7 & 41.1 $\pm$ 5.2 & 13.8 $\pm$ 0.7 \\
Letta/MemGPT\textsubscript{less search} \ifpreprint\citep{packer2023memgpt}\fi & \xmark & 71.0 $\pm$ 0.3 & 81.1 $\pm$ 0.4 & 61.5 $\pm$ 1.1 & 66.7 $\pm$ 1.2 & 45.1 $\pm$ 3.4 \\
Letta/MemGPT\textsubscript{more search} & \xmark & \underline{72.8} $\pm$ 0.2 & \textbf{83.3} $\pm$ 0.4 & \underline{66.6} $\pm$ 1.2 & \underline{71.4} $\pm$ 1.2 & 93.8 $\pm$ 9.9 \\
\midrule
\oursimpl{} \lstinline|invoke| (ours) & \checkmark & \textbf{74.1} $\pm$ 0.6 & \textbf{83.3} $\pm$ 0.7 & \textbf{72.5} $\pm$ 1.3 & \textbf{76.0} $\pm$ 1.4 & 15.6 $\pm$ 0.6 \\
\bottomrule
\end{tabular}
\\
\footnotesize{
*\,See \Cref{app:prompt_only} for our definition of a prompt-only setup with a generic agent loop \\
}
\end{center}
\vskip -0.1in
\end{table}
\else
\begin{figure}[tb]
    \centering
    \vspace{-0.1in}
    \includegraphics[width=0.7\linewidth]{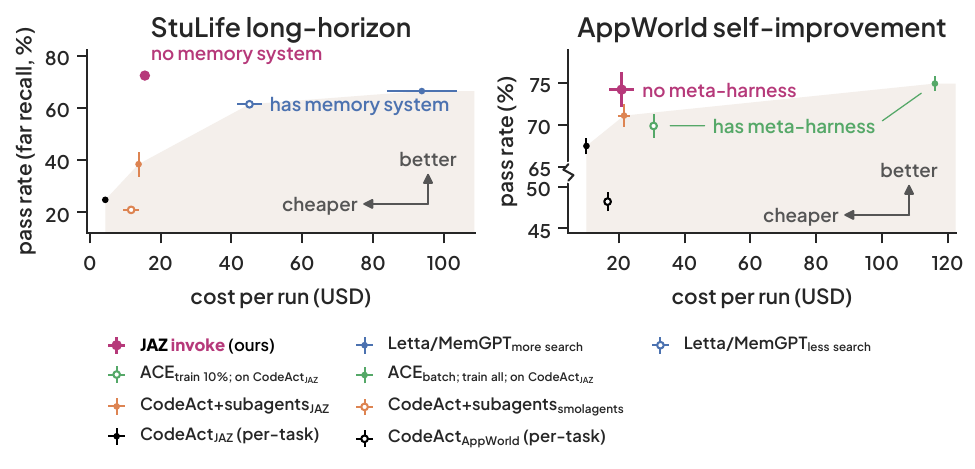}
    \vspace{-0.1in}
    \caption{\oursimpl{} with only prompting can outperform specialized systems at a lower cost.}
    \vspace{-0.2in}
    \label{fig:placeholder}
\end{figure}
\fi

\paragraph{Experiments.}
We evaluate on the long-horizon benchmark
StuLife \citep{cai2026stulife}, which contains an ordered sequence of 1284
tasks that simulate a variety of activities a college student does over the course
of an academic year, many of which require recalling information given in prior tasks.
A typical episode takes 7000--8000 environment interactions, and a typical code-mode agent requires
3000--5000 LLM calls.
Among the 1284 tasks, 939 are graded, of which 207 require ``far recall'',
which we define to be recall of information delivered in a previous task separated by over 50 tasks.

We evaluate \oursimpl{} \lstinline|invoke| and multiple baselines using
GPT-5.4 nano as the model.
Since \oursimpl{} \lstinline|invoke| is a general agent harness,
our primary baseline is CodeAct+subagents, also a general agent harness.
The CodeAct (per-task) baseline runs CodeAct on each individual StuLife
task with no cross-task memory persistence and thus serves as a no-memory baseline.
We also compare with a domain-specific harness: Letta v0.16.8, a conversation agent
with memory. Letta v0.16.8 is the latest version of Letta that is based on
the MemGPT memory architecture \citep{packer2023memgpt}, so we will also refer to it
as ``Letta/MemGPT''.
See \Cref{app:long} for experiment details.
Evaluation results are reported in \Cref{tab:long_horizon}.
An analysis is given below, with details provided in \Cref{app:stulife_traces}.

\paragraph{Analysis.}
To understand where \oursimpl{} \lstinline|invoke| has an advantage
over CodeAct+subagents and Letta/MemGPT on tasks requiring long-range recall,
we analyzed task \#1244/1284, which \lstinline|invoke| solved in 5 of 6 runs, CodeAct+subagents in 0 of 12, and
Letta/MemGPT in 4 of 12.
This task is a final exam multiple-choice question. Each of the four options describes
applying a fictional protocol, and the agent has to choose the correct option by recalling the lectures that taught those protocols,
then apply them in the scenarios given in the options.
The lecture that taught the protocol involved in the correct option (D)
was introduced in task \#953/1284, 291 tasks ago.

In a representative \oursimpl{} \lstinline|invoke| run, when
the agent reached this task, it was already at
recursion depth 66 after having performed tail-recursive delegation 65 times.\footnote{
Recursion depth is defined as 0 in user code, 1 in the top-level agent,
2 in the first level of subagents, etc.
}
However, every delegation preserves the full history completely
through the \lstinline|prev_history|, which has accumulated every prior
agent's \lstinline|__history__|.
The agent spent one turn searching all protocol names in its
\lstinline|prev_history|.
All the relevant lectures were correctly retrieved,
and the agent answered correctly on its next turn.

CodeAct+subagents did not search on this task in any of its runs, as important information
was lost across subagent delegation boundaries.
For example, the user prompt explicitly instructs the agent to search for lecture content
on quiz/exam questions as they test fictional knowledge,
and to maintain its own REPL history to compensate for CodeAct's lack of
the \lstinline|__history__| variable.
However, without access to the user prompt as a variable in the REPL, every agent
has to copy its user prompt into the subagent. Despite the prompt explicitly telling it to
copy the prompt verbatim, it was still lossy:
after multiple delegations, both instructions were lost or heavily degraded.
As a result, in nearly every run, the agent eventually stopped maintaining its history,
so it was no longer able to perform long-range recall in later tasks.

\begin{promptbox}
\textbf{Takeaway 1:} \textit{(long-horizon version)}
When \textit{everything} the agent sees in its context is in REPL variables,
the agent can pass them by reference to a subagent in scenarios that
need it (e.g., tail-recursive delegation).
This is more reliable than copying their contents by hand.
\end{promptbox}

Letta/MemGPT used its \lstinline|conversation_search| tool
to search through its conversation history for the protocol names mentioned in the question.
Using a combination of keyword search and vector search,
the tool returned the top-ranked hits, which included noise such as the quiz question itself
and earlier messages about similarly named but different protocols.
In this scenario where the agent needs exact substring matching and a filter
to exclude the messages that are in context,
the only tool the agent had (\lstinline|conversation_search|) did not easily support either.

\begin{promptbox}
\textbf{Takeaway 2:} \textit{(long-horizon version)}
Domain-specific harnesses encode assumptions that make them work well in many situations,
but they become rigid in environments that break those assumptions.
Instead of a search tool over a large object,
having full programmatic access to its raw interface
can recover the flexibility needed in such environments.
\end{promptbox}

\ifpreprint
\else

\fi

\subsection{Continual self-improvement} \label{sec:csi}



We study \textit{continual self-improvement (CSI)}, where a system
improves itself over a given ordered sequence of tasks.
%
One form of self-improving system separates a \textit{meta-agent}
from a \textit{solver agent}, where the meta-agent optimizes various aspects
of the solver agent. Targets of optimization can include the
prompt \citep{zhang2025ace,dynamiccheatsheet,agrawal2025gepa}, 
executable skills \citep{wang2024voyager},
and even the agent's full
source code \citep{lee2026metaharness}.

In \oursimpl{}, the top-level \lstinline|invoke| has control over all inputs to pass 
into a sub-invoke and can thus act as a meta-agent by optimizing
those inputs over a sequence of tasks.
A typical optimization iteration involves constructing inputs
such as the solver agent's prompt and skills (meta-authored tools), running them on a task,
and inspecting the results, as described by the following pseudocode:
\begin{lstlisting}[basicstyle=\ttfamily\small, upquote=True]
instructions = "... When done, return `(answer, __history__)`"

def tool(...): ...  # function that calls base environment tools

answer, trajectory = invoke(
    task=get_next_task(),
    instructions=instructions,
    tool=tool,
)
eval_report = complete_task(answer)
print(eval_report, trajectory)
\end{lstlisting}
The prompt used to guide the top-level agent to conduct continual self-improvement
is given in \Cref{app:csi_user}. It describes the high-level continual self-improvement
workflow and provides information about \oursimpl{} needed for proper implementation
of self-improvement. The prompt does not include any code examples or templates.

Since our CodeAct baseline runs on each individual task without subagents,
our CodeAct+subagents and \oursimpl{} \lstinline|invoke|
implementations cap recursion depth to 2 to reduce confounds
(i.e., solver subagents are not given sub-subagents).

\ifpreprint
\begin{table}[tb]
\small
\centering
\caption{Results on the full \texttt{test-challenge}
split of AppWorld \citep{trivedi2024appworld}.
Methods marked ``per-task'' solve each AppWorld task
individually with no continual learning across tasks.
The solver agent in all methods uses GPT-5.4 nano (high)
and the meta-agent in CSI methods uses GPT-5.4 (high).
Per-task methods report a 3-run mean and combined per-task SEM,
and all other methods report a 6-run mean and SEM to account for higher variance.
See \Cref{tab:csi_full} for all individual data points.
}
\vspace{0.05in}
\label{tab:csi}
\setlength{\tabcolsep}{5.5pt}
\begin{tabular}{rcccccc}
\toprule
 & & \multicolumn{5}{c}{AppWorld (\texttt{test-challenge})} \\
\cmidrule(lr){3-7}
 & \smash{\scriptsize\shortstack{prompt-\\only?\,*}} & TGC (\%) & SGC (\%) & Cost (\$) & Meta \$ & Solver \$ \\
\midrule
CodeAct\textsubscript{AppWorld \ifpreprint\citep{trivedi2024appworld}\fi} \scriptsize{(per-task)} & \xmark\,$^\dagger$ & 48.2 $\pm$ 1.1 & 20.4 $\pm$ 1.6 & \underline{16.6} $\pm$ 0.2 & --- & 16.6 $\pm$ 0.2 \\
CodeAct\textsubscript{\oursimpl{}} \ifpreprint\citep{wang2024codeact} \fi \scriptsize{(per-task)} & \checkmark & 67.5 $\pm$ 1.0 & 43.4 $\pm$ 1.7 & \textbf{10.2} $\pm$ 0.1 & --- & \textbf{10.2} $\pm$ 0.1 \\
CodeAct+subagents\textsubscript{\oursimpl{}} \ifpreprint\citep{zhang2025rlm}\fi & \checkmark & 71.1 $\pm$ 1.3 & 47.6 $\pm$ 2.1 & 21.6 $\pm$ 1.8 & \textbf{7.3} $\pm$ 1.7 & 14.3 $\pm$ 0.9 \\
ACE\textsubscript{train 10\%; on CodeAct\textsubscript{\oursimpl{}}} \ifpreprint\citep{zhang2025ace}\fi & \xmark & 69.9 $\pm$ 1.4 & 47.1 $\pm$ 1.3 & 30.6 $\pm$ 0.5 & 17.1 $\pm$ 0.3 & 13.5 $\pm$ 0.3 \\
ACE\textsubscript{batch; train all; on CodeAct\textsubscript{\oursimpl{}}} & \xmark & \textbf{74.9} $\pm$ 0.9 & \textbf{52.2} $\pm$ 1.4 & 116.1 $\pm$ 0.9 & 102.2 $\pm$ 0.7 & 13.9 $\pm$ 0.3 \\
\midrule
\oursimpl{} \lstinline|invoke| (ours) & \checkmark & \underline{74.2} $\pm$ 2.1 & \underline{51.1} $\pm$ 3.7 & 20.9 $\pm$ 3.7 & \underline{9.9} $\pm$ 3.8 & \underline{10.9} $\pm$ 0.6 \\
\bottomrule
\end{tabular}
\\
\footnotesize{
*\,See \Cref{app:prompt_only} for our definition of a minimal, prompt-only setup with a generic agent loop
\\
$^\dagger$\,The official CodeAct baseline uses a prompt that contains AppWorld-specific workflow guidance.
}
\vskip -0.1in
\end{table}

\fi

\paragraph{Experiments.}
We evaluate \oursimpl{} \lstinline|invoke|'s ability to
continually self-improve across the full sequence of 417 tasks
from the \texttt{test-challenge} split of the AppWorld benchmark \citep{trivedi2024appworld}.
The task ordering was shuffled with seed 42.
AppWorld is an environment that simulates the API interfaces of 9 apps
people commonly use on their phone or computer, and tasks require
the agent to use those APIs to complete tasks for the user.
In our continual self-improvement setup, the full set of tasks is given to the agent
in a fixed order, with full test feedback from each task so that the agent
can learn from its mistakes.

We compare \oursimpl{} \lstinline|invoke| with an equivalent minimal prompt-only setup
of CodeAct+subagents, as well as Agentic Context Engineering (ACE)
\citep{zhang2025ace}, a specialized harness for continual self-improvement.
We also compare with baselines that solve each benchmark task individually:
the benchmark's official CodeAct baseline\footnote{Known as the ``official ReAct'' baseline
with a REPL tool, which is equivalent to CodeAct.},
as well as our own implementation of CodeAct.
These represent baseline methods that do not use test feedback
to improve across tasks.

\Cref{tab:csi} reports the mean and standard error of our evaluation results.
\Cref{tab:csi_full} lists all individual data points, and additionally reports
the median with its 78\% confidence interval for self-improving methods.
We find that \oursimpl{} \lstinline|invoke|
outperforms other methods that have a similar cost and is much cheaper than methods that have a similar performance.

\ifpreprint
\else

\fi

\paragraph{Analysis.}
We qualitatively compare the continual self-improvement behavior of prompt-only \oursimpl{} \lstinline|invoke|,
prompt-only CodeAct+subagents, and the specialized self-improvement harness ACE \citep{zhang2025ace}.
Excerpts from the traces of both prompt-only methods are given in \Cref{app:appworld_traces}.

\oursimpl{} \lstinline|invoke|'s top-level \lstinline|invoke| acted as a meta-agent that launched
subagents to solve individual tasks. It dispatched tasks in batches to subagents. Between batches, it
inspected metrics, the grader's failure messages, and selected parts of failed subagents' trajectories,
updated inputs to subagents (prompts in every run, and meta-authored tools in 5 of the 6 runs),
and adaptively updated the batch size. To access subagent trajectories,
the meta-agent simply had to tell the subagent to return its \lstinline|__history__| variable
available in its REPL.

CodeAct+subagents followed the same workflow as \oursimpl{} \lstinline|invoke|, but it had to get around
the limitation that subagents didn't have access to their own REPL histories.
So we prompted the meta-agent to tell each subagent to return its REPL history.
Depending largely on the prompt the meta-agent wrote, the subagent either hard-coded the history
in its final return statement --- often reducing it to a single entry --- or incrementally built it by appending to it in each turn.
Both approaches were lossy, and the additional overhead contributed to more output tokens 
and higher cost in solver agents.

\begin{promptbox}
\textbf{Takeaway 1:} \textit{(self-improvement version)}
When \textit{everything} a subagent sees in its context is in REPL variables,
the subagent can pass them by reference to its caller in scenarios that
need it (e.g., for caller observability into subagent behavior).
This is more reliable than explicitly building their contents.
\end{promptbox}

ACE is a human-designed self-improvement loop that optimizes the solver agent's prompt
after every task it attempts. It uses a \textit{reflector} LLM call to reflect on the solver agent's
trajectory, and then a \textit{curator} to distill findings into bullet points that are added
to the \textit{playbook} that forms part of the solver agent's prompt.
The ACE reflector and curator were very costly, costing 10--31x more than \oursimpl{} \lstinline|invoke|'s top-level agent to train across all tasks depending on batching
(\Cref{tab:csi_full}).
The cost efficiency of \oursimpl{} \lstinline|invoke| comes from breaking free of
the rigid workflow prescribed by ACE.
In \oursimpl{} \lstinline|invoke|, the meta-agent has freedom on how to batch tasks
and selectively reads parts of trajectories,
thus saving its context from being filled with the full trajectory of every subagent.

Another limitation of ACE is that it only optimizes the prompt, whereas the \oursimpl{} \lstinline|invoke|
meta-agent optimizes inputs to \lstinline|invoke| generally, which can also include meta-authored tools.

\begin{promptbox}
\textbf{Takeaway 2:} \textit{(self-improvement version)}
Hand-designed workflows encode assumptions that make them work well in many situations,
but they become rigid in environments that break those assumptions.
Providing a strong LLM the full expressivity of code and the \lstinline|invoke| primitive
can recover the flexibility needed in such environments.
\end{promptbox}

\ifttcl\else\ifiab\else
\section{Related work} \label{sec:related}

\paragraph{Agent loop paradigms.}
Most modern LLM agents are built around an \emph{agent loop} in which the model interacts repeatedly with an environment through actions and observations. In traditional tool-calling or ReAct-style agents~\citep{yao2022react},
each action invokes a \textit{tool} with hard-coded inputs.
More recent \emph{code-mode} or \emph{CodeAct-style} agents instead place the model inside
a code environment (REPL), where tools and subagents are now ordinary functions that code can reference~\citep{wang2024codeact,zhang2025rlm}.
\oursimpl{}'s \lstinline|invoke| primitive implements a
code-mode agent loop where all inputs to the agent (including task instructions) and the REPL history
--- together, everything the LLM sees --- are variables in the REPL.


\paragraph{Minimal agent harnesses.}
LLM agents have seen a broad trend towards simpler designs as models become stronger.
Early systems often had a human-engineered workflow in the form of a graph
or program \citep{ibrahimzada2025alphatrans,lu2024aiscientist,shetty2024syzygy,wang2023hypothesis,zelikman2023parsel}.
They gradually gave way to agent loops \citep{yao2022react,wang2024codeact,anthropic2026claudecode,wang2025openhands}
where the LLM acts in an environment through an interface.
Minimal but expressive interfaces have been found to be effective.
For example, mini-SWE-agent \citep{lieret2025minisweagent}
showed that giving an LLM \lstinline|bash| is sufficient for software engineering,
and RLMs \citep{zhang2025rlm} showed that an LLM with a Python REPL with
recursive subagents can perform long-context tasks.
Continuing in this direction of removing complexity from agent harnesses,
\oursimpl{} demonstrates that an LLM with a Python REPL
where everything it sees is a variable can handle agentic tasks requiring long-range recall,
as well as continual self-improvement.

\paragraph{LLM agent frameworks.}
Frameworks for LLM agents have tracked the evolution towards simpler harnesses
with more expressive agent loops. Many of the earlier agent frameworks
\citep{chase2023langchain,campos2025langgraph,khattab2024dspy,li2025encompass}
aimed to help developers build graphs or programs that hard-code a workflow.
More recent ones \citep{langchain2025v1,roucher2025smolagents,t2023claudeagentsdk,mehta2023openaiagentssdk,colvin2024pydanticai,moura2023crewai}
are centered around an agent loop, where hooks extend and modify
the agent. The \oursimpl{} framework is similar; its main contribution is
a more expressive agent loop with a simpler and more capable out-of-the-box
default agent.

\paragraph{Multi-agent systems.}

Multi-agent systems for task decomposition have been built
for solving complex long-horizon tasks~\citep{hong2023metagpt, qian2024chatdev, chen2024divide, zelikman2023parsel, light2025disc}.
Traditionally involving significant hand design, such as
subagent roles and specific orchestration patterns,
more recent multi-agent systems \citep{anthropic2026claudecode,zhang2025rlm} take a more minimal approach involving generalist subagents
exposed as regular tools.
While traditionally subagents have been used for well-defined subtasks,
we demonstrate that they are also useful for long workflows without well-defined subtasks,
and can also be used as the solver agent in a self-improving system with a meta-agent/solver-agent split.

\paragraph{Compaction and memory systems.}

For non-decomposable long-horizon workflows such as long conversations, the predominant
approach has been \textit{compaction} and \textit{memory systems}.
Compaction methods include simple textual summarization \citep{wang2025compact,packer2023memgpt}, and recent work
has also explored applying KV cache compaction techniques to long-horizon agentic workflows
\citep{liu2026kvcompact}. To enable retrieval of previous conversation context that has been
lost due to compaction, memory systems were developed that provide the agent an interface
to query conversation history through keyword or embedding-based search \citep{packer2023memgpt,chhikara2025mem0}.
In \oursimpl{}, we showed that full programmatic access to a variable
\lstinline|__history__| is sufficient for memory retrieval.

\paragraph{Self-improving agentic systems.}
Self-improving systems take a variety of forms, including agents that modify their own source code
\citep{zhang2026dgm,zelikman2024stop,robeyns2025sica,yin2025godelagent}
and systems involving a meta-agent that optimizes a target agent
\citep{khattab2024dspy,agrawal2025gepa,hu2025adas,zhang2025ace,lee2026metaharness}.
Aspects of the target agent to be optimized can include the prompt
\citep{khattab2024dspy,agrawal2025gepa}, skills (high-level tools) \citep{wang2024voyager, light2025strategist},
and entire agent programs \citep{hu2025adas,lee2026metaharness}.
In \oursimpl{}, we demonstrated that the top-level agent and subagent in a multi-agent system can
act as the optimizer meta-agent and the optimized solver agent in self-improvement,
optimizing inputs to the solver agent such as its prompt and tools.

\paragraph{Program synthesis with sketching.}
\oursimpl{} also takes inspiration from program synthesis with sketching
\citep{solar2005sketch,solar2006sketch,solar2013sketch},
where a program with holes has its holes filled in by a synthesizer to satisfy a specification.
Functional \lstinline|invoke| can be viewed as a generator in sketching
whose entire body is a hole
that is synthesized by an LLM at
every invocation.

\fi\fi




\section{Conclusion}
We demonstrated that behaviors commonly attributed to specialized agent harnesses
(e.g., memory and self-improvement) can instead be elicited through prompting,
once the agent loop becomes expressive enough to allow the model to express these behaviors.
This suggests a shift in how LLM agents should be built.
Instead of harness engineering,
the more ``bitter-lesson pilled'' approach may be to train a model within a minimal yet
maximally expressive agent loop,
so that it could learn to construct workflows, tools, and harness components
by itself. A minimal harness with the least inductive bias allows
for the most general-purpose learning, and maximal expressivity gives the agent the
freedom to achieve any desired behavior or capability.

\ifmetaagents
\section*{Responsible Use Statement}

This work demonstrates that a minimal harness offering the model enough freedom ---
allowing it to access parts of its context by reference and orchestrate subagents through code ---
can replicate the behavior of more sophisticated harnesses with manual orchestration.
The potential capabilities of such a system have been demonstrated to be similar to those
of existing harnesses while having a simpler design.
One should follow general responsible usage guidelines for LLM-based agents
and optimizing meta-agents, e.g., one should not use a meta-agent to optimize for misaligned behavior.
\fi

\begin{ack}
This project was funded by the following grants and institutions: Zup/Ita\'u,
MIT CSAIL FinTech AI Initiative, NSF Grant No.~1918839.

\end{ack}

\bibliography{refs}
\bibliographystyle{unsrt}

\newpage
\appendix
\crefalias{section}{appendix}\crefalias{subsection}{appendix}\crefalias{subsubsection}{appendix}
\ifttcl

\else\ifiab

\fi
\fi
\section{Formal definition of \texorpdfstring{\lstinline[basicstyle=\ttfamily\large]|invoke|}{invoke}} \label{sec:formal}

Given a programming language, we augment its syntax with the
\textit{\lstinline|invoke| primitive},
which is a function call where the body of the function is written
dynamically at runtime by a language model ($\mathsf{LM}$).
We write \lstinline|invoke|$(x_1 := e_1, \ldots, x_k := e_k)$
for the functional agent call on \textit{inputs} $x_i := e_i$,
where $x_i$ are variable names and $e_i$ are expressions in the language.
To execute a program in this augmented language,
\lstinline|invoke|$(x_1 := v_1, \ldots, x_k := v_k)$ is evaluated as follows:
\begin{enum}
    \item Query a language model with a prompt $P$ that contains
    a compact string representation
    $P = \langle x_1 := v_1, \ldots, x_k := v_k\rangle$ of the inputs.
    These inputs can include the user prompt, tools, and more generally
    any object in the language allowed as an argument to a function call.
    \item The model returns code $C = \pi(\mathsf{LM}(P))$
    in the same language
    that would be valid as the body of a function with
    input parameters $x_1, \ldots, x_k$.
    Here, $\pi(\cdot)$ parses the code from the model's raw output.
    Note that all the context that the model sees is contained
    within the prompt $P$ that serializes the inputs.
    \item Execute this function on the given inputs.
\end{enum}

Since the language itself includes the \lstinline|invoke| primitive,
the model-generated code may recursively contain further
\lstinline|invoke| calls.

Under this framework, there are three key aspects to design: the language itself,
the \textit{serializer} formatting the inputs into a prompt
(represented by the angle bracket notation ``$\langle \cdot \rangle$''),
and the \textit{model output parser} $\pi(\cdot)$.
A good serializer does not display the raw contents of a large object;
instead it gives the model enough information so that it could extract information needed
by interacting with it.
The model output parser traditionally involves parsing out a code
block from the model's natural language output. However, for simplicity,
it can just be the identity function $\pi(y) = y$,
which was the choice we made in \oursimpl{} ---
the entirety of the model's output is parsed as Python code,
and any natural language prose
(e.g., the agent's plan) is included in leading comments inside the code.

Formally, consider the lambda calculus with constants, a minimal functional programming language
with just constants, variables, function application, and function abstraction.
The following formal definition is easily generalized to other programming languages.

Augmenting this calculus with the \lstinline|invoke| primitive results
in this augmented grammar of expressions (terms):
\begin{equation}
e ::= \underbrace{c \mid x \mid e\ e \mid \lambda x.\,e}_{\mathclap{\text{lambda calculus with constants}}} \mid \text{\lstinline|invoke|}\ (x := e)^+.
\end{equation}
The standard expressions are constants ($c$),
variables ($x$), function calls ($e\ e$), and function definitions ($\lambda x.\,e$).
The additional syntax is the new primitive $\text{\lstinline|invoke|}\ (x := e)^+$.
Here, each $e$ denotes an arbitrary expression
in the language, and $(x := e)^+$ means ``one or more named arguments''.

The big-step operational semantics for call-by-value (CBV) evaluation
become
\begin{equation}
\begin{gathered}
\underbrace{
\infer[\tiny\text{1}]{v \Downarrow v}{}
\qquad
\infer[\tiny\text{2}]{e_1\ e_2 \Downarrow v}{e_1 \Downarrow \lambda x.\,e & e_2 \Downarrow v_2 & e[v_2/x] \Downarrow v}
\qquad
\infer[\tiny\text{3}]{e_1\ e_2 \Downarrow v}{e_1 \Downarrow c & e_2 \Downarrow v_2 & \delta(c, v_2) = v}
}_{\text{lambda calculus with constants}}
\\[0.5em]
\infer[\tiny\text{4}]{\text{\lstinline|invoke|}\ (\vec x := \vec e) \Downarrow v}{\vec e \Downarrow \vec v & \pi(\mathsf{LM}\langle \vec x := \vec v\rangle)[\vec v/\vec x] \Downarrow v}
\end{gathered}
\end{equation}
Here, $e \Downarrow v$ means that expression $e$ evaluates to value $v$,
where the metavariable $v$ ranges over values $v ::= c \mid \lambda x.\,e$.
In rule 4, $\vec x := \vec e$ abbreviates the named arguments $x_1 := e_1, \ldots, x_k := e_k$,
and $\vec e \Downarrow \vec v$ means $e_i \Downarrow v_i$ for each $i$.
The first three rules are standard: values evaluate to themselves (rule 1);
a function call evaluates the function and its argument, then either evaluates the function body
with the parameter substituted with the evaluated argument (rule 2) or,
if the function is a constant, looks up the result through the given $\delta$ (rule 3).
Our addition (rule 4) defines how to evaluate the functional \lstinline|invoke| primitive,
formalizing the notion of filling in the body of a function dynamically at runtime:
the arguments are evaluated to values $v_1, \ldots, v_k$ first, just as in a function call,
and then the generated code is evaluated with these values in place of $x_1, \ldots, x_k$.
$\mathsf{LM}\langle \vec x := \vec v\rangle$
denotes the output of the language model given the \textit{prompt} $\langle \vec x := \vec v\rangle$,
i.e., a string representation of all the inputs.
$\pi$ is the \textit{model output parser} that parses the model's raw output.
The extracted code is the body of a function with parameters $x_1, \ldots, x_k$,
and $[\vec v/\vec x]$ substitutes the argument values, as in rule 2.


\newpage
\section{Additional details of \oursimpl{}} \label{app:method_details}

\subsection{Configuration system of \oursimpl{}} \label{app:config}

A configuration in \oursimpl{} has a simple signature:
\lstinline|Config(llm: BaseLLM, repl: BaseREPL, protocol: BaseProtocol)|.
The \lstinline|llm| component configures the model, the \lstinline|repl| component configures the REPL,
and the \lstinline|protocol| component configures the seam that connects the two,
such as parsing the code out of the model's raw response and formatting the REPL output for the model.
\oursimpl{}'s default protocol treats the full raw response as the code with no parsing,
and truncates the REPL output if it's too long but otherwise applies no additional formatting.
Similar to variables and hooks, a configuration can also be either local or subject to dynamic scoping:
a \lstinline|ConfigOverride(llm=..., repl=..., protocol=...)| passed positionally to \lstinline|invoke|
applies only to that \lstinline|invoke| and not to any of its sub-invokes, whereas
activating it as a context manager (\lstinline|with ConfigOverride(...): invoke(...)|)
applies it to all \lstinline|invoke| calls under the \lstinline|with| block,
including all recursive sub-invokes.

\subsection{Extensibility via hooks in \oursimpl{}} \label{app:extensibility}

In \oursimpl{}, users can define custom hooks by providing functions to call at various points
in the agent's lifecycle. Instead of allowing a hook to arbitrarily modify agent state,
it is only allowed to emit static composable effects as allowed by the particular event.
\begin{itemize}
    \item \lstinline|InvokeEnter|, \lstinline|InvokeSend|, \lstinline|InvokeComplete|, \lstinline|InvokeExit|
    \item \lstinline|LLMQueryEnter|, \lstinline|LLMQuerySend|, \lstinline|LLMQueryComplete|, \lstinline|LLMQueryExit|
    \item \lstinline|LLMQueryRetry| (only fires if an LLM query did not succeed and is retried)
    \item \lstinline|REPLExecEnter|, \lstinline|REPLExecSend|, \lstinline|REPLExecComplete|, \lstinline|REPLExecExit|
\end{itemize}
An event handler at \lstinline|[Operation]Enter| sees the inputs to \lstinline|[Operation]| and can emit effects that modify those inputs.
An event handler at \lstinline|[Operation]Send| sees the input modifications and the resulting modified inputs, and can
emit an effect that replaces calling an \lstinline|[Operation]| with supplying its output.
An event handler at \lstinline|[Operation]Complete| sees the output and can emit effects that modify the output.
In addition, at every event other than \lstinline|InvokeExit| and \lstinline|LLMQueryRetry|,
the handler may emit an effect that aborts the \lstinline|invoke| it fires in.

Every \lstinline|[Operation]Enter| opens a span that always closes with \lstinline|[Operation]Exit|,
carrying its outcome (\lstinline|Completed|/\lstinline|Aborted|/\lstinline|Failed|).
\lstinline|[Operation]Send| and \lstinline|[Operation]Complete| are conditional:
an abort or failure skips to \lstinline|[Operation]Exit|.

When multiple hooks are active at once,
effects that modify event fields are \textit{not} applied as they are emitted, as that would cause
different hooks to potentially see different event fields in a way that
depends on the order in which the event is dispatched to all the active handlers.
Instead, the event is dispatched to all handlers at once, and all effects are collected
and composed with a composition rule that minimally depends on composition order.
Conflicting writes that cannot be resolved are refused with an error,
and then the final resolved modifications are all applied at once.

Hooks can also communicate via a global message board called the \textit{blackboard},
reading its contents from the event and writing by emitting an effect.

All handlers also have a view of ambient state:
all active hooks, all scoped variables (\Cref{sec:jaz}), the active configuration (\Cref{app:config}),
and the blackboard.

All built-in hooks provided by \oursimpl{} are built using the hook interface
we've described in this section.

\newpage
\section{Experiment details} \label{app:exp}

\subsection{Evaluation protocol} \label{app:eval_protocol}

To ensure fair comparison, accurate measurements, and proper control of confounds,
we adopt the following protocol in our evaluations.

\paragraph{Task/method decoupling.}
The hyperparameter with the most degrees of freedom is the prompt.
If each (method, task) pair gets its own prompt, there is a high risk of overfitting the method to the task,
and different methods under comparison may receive different degrees of overfitting that
confound the comparison.
This is especially dangerous with self-improvement, where allowing a self-improvement method's
meta-prompt to include task-specific information licenses hard-coding aspects of the optimal solver
prompt and skills in the meta-prompt instead of forcing the meta-agent to learn these on its own,
which would no longer isolate the effect of self-improvement itself.

To mitigate these risks, our experiments split the user prompt into two disjoint parts:
\lstinline|instructions| and \lstinline|guidance|.
The environment-provided \lstinline|instructions| prompt is method-agnostic --- it provides the task
instructions, describes the environment, and optionally provides workflow guidance as long as it is
method-agnostic. In self-improvement experiments,
to isolate the ability to self-improve starting from a minimal seed,
\lstinline|instructions| contains only the task instructions and environment description and
provides no workflow guidance at all. On the other hand, the method-provided \lstinline|guidance| is task-agnostic
--- it provides workflow guidance as appropriate for the method, as long as it is applicable
to all tasks in the domain. In \oursimpl{}, the prompt components \lstinline|instructions| and \lstinline|guidance|
are passed as separate variables to \oursimpl{} \lstinline|invoke|.

Note that we consider ``[agent loop] long-horizon'' and ``[agent loop] self-improvement''
to be separate methods that can provide their own prompts, but the prompt must still be task-agnostic.
Thus, the long-horizon prompt (\Cref{app:long_horizon_prompt})
and the self-improvement prompt (\Cref{app:csi_user})
are different prompts, but the long-horizon prompt is generally applicable to long-horizon environments
with no StuLife-specific information, and the self-improvement prompt is generally applicable
to self-improvement with no AppWorld-specific information.
When the official implementation of a baseline (e.g., ACE \citep{zhang2025ace}) violates this rule
(\Cref{app:ace_setup}), we make the minimal modification so that it conforms.

\paragraph{Necessary differences between method implementations are implemented.}
A \lstinline|guidance| prompt written for one method is typically not applicable to another method
and thus must be adapted. For example, compared to \oursimpl{} \lstinline|invoke|,
CodeAct+subagents does not expose the user prompt components (\lstinline|instructions| and \lstinline|guidance|)
as variables in the REPL and does not automatically maintain a \lstinline|__history__| variable in the REPL.
Thus, for the long-horizon environment, we adapt the \lstinline|guidance| prompt to ask the agent to maintain
its history by itself, and we adapt the context window warning to ask the agent to hard-code
the contents of \lstinline|instructions| and \lstinline|guidance| instead of referring to their variables.
Similarly, for self-improvement, we adapt the \lstinline|guidance| prompt to ask the meta-agent
to ask the solver agent to maintain its history by itself so that it could return it at the end.

\paragraph{Unnecessary differences between method implementations are minimized.}
When two methods each have tunable hyperparameters --- especially the prompt, but also the configuration
--- those that can be kept the same across the two methods without introducing bias towards one method
or the other are kept the same. For example, \oursimpl{} truncates inputs to 50,000 characters by default,
so in the smolagents implementation of CodeAct+subagents, the context window warning asks the agent
to wrap its history list in an object whose custom \lstinline|repr| truncates to 50,000 characters.
For prompts, the \oursimpl{} \lstinline|invoke| prompt and the CodeAct+subagents\textsubscript{\oursimpl{}}
prompt are byte-identical everywhere other than in places where a difference is necessary.

\paragraph{Cost measurement.} For accurate cost measurement, we controlled for systematic errors in cache hit rate in two ways.
First, we set a distinct prompt cache key for each run.
Second, we allowed at most 3 parallel runs at any moment under the same OpenAI organization.
(Multiple OpenAI organization accounts were created to run more than 3 experiments in parallel.)
To estimate the systematic uncertainty due to uncontrollable factors in the state of
OpenAI's server-side cache, we ran $n$ parallel agents on a simple toy environment
for $n=1, 3, 4, 5$, both with and without another agent running on a different environment at the same time.
We calculated the mean and its standard error of the cache hit rate.
We found a notable decrease in cache hit rate of over 5\% from $n = 4$ to $n = 5$,
while changes in cache hit rate for $n \leq 4$ were within 2\%.
We therefore fix $n = 3$ for all our experiments, and estimate the systematic error in
resultant cost estimates to be less than 2\%.
Experiments that involved 6 independent runs were run in 2 batches of 3.

\paragraph{Rules for the prompt-only setup with a generic agent loop.} \phantomsection\label{app:prompt_only}%

The ``minimal'' setup we apply to all general agent loop methods in our
experiments applies the following restrictions to all methods equally:
\begin{items}
    \item The only tools are those provided by the benchmark's
    environment itself. More generally, the agent is not allowed to access anything
    other than the given environment, such as the file system or the Internet.
    \item There is no loop-external logic that modifies the loop itself
    (e.g., truncating the conversation history, injecting variables in the REPL).
    However, \textit{monitoring} is allowed:
    budget control, step limits, recursion limits, completion guards, and context window warnings.
    We also allow model configuration, i.e., the root agent may use a different
    model from subagents.
    \item The system prompt --- available to the root agent
    as well as subagents --- contains agent loop mechanics
    and tool docstrings.
    \item The user prompt --- passed to the root agent only
    --- contains a method-agnostic task prompt supplied by the environment
    and a task-agnostic guidance prompt supplied by the method.
\end{items}
The details of implementing a general agent loop may depend on the domain
(``long-horizon'' or ``self-improvement''), but must be task-agnostic.
For example, the method's contribution to the user prompt and context window warning must be applicable to
``long-horizon'' in general, or ``self-improvement'' in general, with StuLife-
or AppWorld-specific information provided only by the environment in a method-agnostic manner.

\subsection{Long-horizon: StuLife} \label{app:long}

\subsubsection{Environment details}

The StuLife repository provides a JSON file containing the 1284 tasks of the environment.
We implement an environment that delivers these tasks and exposes a modified set of tools
to the agent. We reuse the official grader for binary scoring and implement our own partial score metric.
Our modified set of tools removes tools that are not needed for any tasks, and replaces a few tools
that are always used the same way with a single tool for the same purpose (see code for details).
The original tools also only output unstructured text and never raise on invalid input,
so they are wrapped to take structured inputs, return structured outputs, and perform
input validation to be friendly to code-mode agents. Their docstrings are also updated accordingly,
and code examples are included in docstrings to improve clarity.

\subsubsection{Method details}

For our long-horizon experiments with \oursimpl{}, we have two prompts:
one is the warning emitted when the LLM's context window is 70\% full,
injected with the \lstinline|ContextWindowWarning| hook,
and the other is part of the regular user prompt given to the top-level agent.
The prompt injected by the \lstinline|ContextWindowWarning| is given in
\Cref{app:context_window_warning}.
The prompt shown to the top-level \lstinline|invoke| is provided inside
a regular input named \lstinline|guidance| and its contents are given in
\Cref{app:long_horizon_user}.

We implement CodeAct+subagents as a minimal modification of \oursimpl{} to minimize confounds.
We apply the \lstinline|CodeAct| hook that removes \lstinline|__history__| and the user prompt(s)
from the REPL of every \lstinline|invoke|, and make the minimal modification to the
long-horizon \lstinline|guidance| and \lstinline|ContextWindowWarning| prompt to accommodate the removal.
In particular, the \lstinline|guidance| prompt additionally shows the agent how to maintain its own
REPL history by appending intermediate outputs to a list \lstinline|output_history|,
and the history search section is updated to use this variable.
The context window warning text is adapted to use this \lstinline|output_history| variable,
and instead of showing the user prompt variables (\lstinline|instructions| and \lstinline|guidance|)
being passed to the subagent by reference, the warning text demonstrates that they should be 
hard-coded as literal strings copied from the agent's context.

For Letta/MemGPT \citep{packer2023memgpt}, we use the Python Letta server
of the latest version of Letta that is based on the MemGPT memory architecture,
namely, v0.16.8.
Letta v0.16.8 supports parallel tool calling, so we enable it to more closely
match the setup in \oursimpl{} where a single step can include multiple tool calls.
Letta v0.16.8 provides two search backends.
The SQL backend caused most search results to turn up empty:
it applies exact substring search to the \textit{entire} search query,
yet the agent issues search queries that are more appropriate for conventional search engines
and thus frequently do not appear as an exact substring of anything in the history.
We thus used Letta v0.16.8's more powerful search backend, Turbopuffer, which implements
hybrid search, combining BM25 keyword-based search and vector-embedding-based search.
We had to modify the environment to deliver each task as a user message as opposed to
letting the agent call a tool to retrieve the task, as tool outputs are not stored in Letta v0.16.8's
searchable history. We tried multiple prompts for Letta v0.16.8, including the empty prompt and variations of
\oursimpl{}'s \lstinline|guidance| prompt adapted to Letta v0.16.8.
We found that all but one of the prompts consistently resulted in an unrecoverable infinite loop
where the agent is stuck on an empty-description task and searches for the task description forever.
We report the results of the one prompt that worked as Letta/MemGPT\textsubscript{less search},
as this prompt also led the agent to search less.
We also ran Letta v0.16.8 with the empty prompt and mitigated the infinite loop issue
by forcing the agent to move on whenever it has spent too many rounds on the same task.
These results are reported as Letta/MemGPT\textsubscript{more search}.

For the CodeAct per-task baseline, our setup is similar to that of CodeAct+subagents,
except each task gets its own CodeAct agent. It is implemented using the \lstinline|CodeAct|
hook on \oursimpl{} \lstinline|invoke|, and the \lstinline|RecursionLimit(max_depth=1)| hook is used
to disable subagents.

The smolagents \citep{roucher2025smolagents} implementation of CodeAct+subagents mirrors that of CodeAct+subagents\textsubscript{\oursimpl{}}.
The managed agents feature is used for subagents, and prompts are adapted to smolagents.
Various workarounds were needed due to limitations in smolagents:
\begin{items}
    \item The \lstinline|additional_args| that seed the subagent's REPL display their \lstinline|repr|
    with no truncation, so the agent is instructed to wrap its history in a custom object
    with a \lstinline|repr| that truncates.
    \item The restricted Python interpreter that smolagents implements for its Python REPL resolves
    a name not by its presence in the namespace, but by fuzzy matching. Our prompts use \lstinline|try|/\lstinline|except| blocks
    to check whether a variable is already defined and define it if not.
    Because of the fuzzy matching, these blocks are ordered so that no such variable fuzzy-resolves to another
    when checking whether it's already defined inside the \lstinline|try| block.
    They catch \lstinline|Exception| because an undefined name
    raises the interpreter's own custom error rather than \lstinline|NameError|.
\end{items}
We noticed other issues in smolagents that had no obvious workaround.
We did not fix them as that would modify the smolagents implementation beyond the tunable surface.
For example, we did not fix the issue in smolagents' Python interpreter where certain functions
from the \lstinline|math| module (e.g., \lstinline|log|) are hard-coded to resolve to the math function,
even when the agent redefines it (e.g., a \lstinline|log| helper function that stores output to the
history and prints).
Separately, we kept smolagents' default system prompt, which calls subagents ``team
members'' and incorrectly says ``this team member is a real human''.
This amplified CodeAct+subagents' underperformance relative to \oursimpl{} \lstinline|invoke|
and was a reason it performed worse than \oursimpl{}'s reimplementation of CodeAct+subagents.
More specifically, once its user prompt had been lost across delegation boundaries,
the agent started launching a subagent whenever it was missing information, perhaps under the
false impression that this ``human team member'' could go find the information for the agent.
Although the subagent's role description we wrote for smolagents clarifies that
it has no access to humans, the effect was minimal.

We chose smolagents instead of RLMs \citep{zhang2025rlm} as our
``existing CodeAct+subagents implementation'' baseline, despite the fact that the RLM
popularized this agent loop paradigm.
The reason was that smolagents is more general. While the RLM implementation restricts
the non-prompt inputs (inputs that become Python REPL variables) to a single string or
JSON-serializable message dictionary or list,
smolagents allows for arbitrary Python objects as inputs.
RLM's implementation in the DSPy framework \citep{khattab2024dspy} would have been
a potential alternative, and it does support arbitrary JSON-serializable objects as
inputs to the agent. However, it did not support recursive subagent calls when we started
our experiments in August 2026 (DSPy 3.3.1 \citep{miller2026dspyrlm})
and was therefore unable to use subagents to conduct our long-horizon and
self-improvement workflows.

\subsection{Continual self-improvement} \label{app:csi}

\subsubsection{Environment details}

The task sequence consists of the full set of 417 tasks from the \lstinline|test-challenge| split
of AppWorld. We randomly permute the task sequence with a fixed seed (seed 42).

AppWorld does not expose its environment \lstinline|apis| directly as methods a user can call,
but instead only indirectly through a Python REPL interface provided by the benchmark authors.
Technical incompatibilities make it difficult to directly replace the \lstinline|invoke|
Python REPL with AppWorld's REPL. Instead, we construct an explicit \lstinline|apis| object
that mirrors the \lstinline|apis| object in AppWorld's REPL. A method call
is resolved by writing the corresponding single line of code that gets executed by AppWorld's REPL,
and the result is retrieved from the namespace of AppWorld's REPL and returned.

AppWorld provides prompts used in its baseline evaluations, but they mix the environment description,
task instructions, and workflow guidance into one prompt. We removed workflow guidance and separated
the remaining content, placing the environment description into the docstring of the environment object
\lstinline|apis| and task instructions into the \lstinline|instructions| component of the user prompt.
We found that the original description of the task submission format had an ambiguity that caused a drop of about 15 percentage points
in the performance of the baseline (\Cref{app:appworld_official_baseline}), so we made sure our
\lstinline|instructions| prompt was clear and complete and exhibited no ambiguities.

\subsubsection{Method details} \label{app:csi_method}

In \oursimpl{} \lstinline|invoke|, we provide a prompt that guides the root \lstinline|invoke|
to perform continual self-improvement. The prompt is provided as a regular
input \lstinline|guidance| to the root \lstinline|invoke|.
Its contents are given in \Cref{app:csi_user}.
We use a scoped (context manager) \lstinline|ConfigOverride| to set
the model to GPT-5.4 nano, but use a local \lstinline|ConfigOverride|
at the top-level to selectively override its own model to GPT-5.4.
We do not set a \lstinline|ContextWindowWarning| hook as the top-level agent
is able to process all tasks within its context window.
We use a \lstinline|RecursionLimit| hook to cap the recursion depth to 2
so that the solver agent doesn't get access to subagents. This ensures that the measured difference
relative to the non-self-improving CodeAct baseline is attributable to self-improvement
alone and is not confounded by solver agents' additional access to subagents.

We implement CodeAct+subagents as a minimal modification of \oursimpl{} to minimize confounds.
We apply the \lstinline|CodeAct| hook that removes \lstinline|__history__| and the user prompt(s)
from the REPL of every \lstinline|invoke|, and make the minimal modification to the
self-improvement \lstinline|guidance| prompt to accommodate the removal.
In particular, the meta-agent is no longer told to ask the solver agent to return its \lstinline|__history__|.
Instead, the meta-agent is told to ``return its REPL history as a list
where each entry holds the code it ran, the output the code printed, and any exception
that occurred''.

\phantomsection\label{app:ace_setup}%
In ACE \citep{zhang2025ace}, the original paper reports that results are insensitive
to the bullet dedup threshold, tested with the values $\{0.5, 0.7, 0.9\}$.
We set the dedup threshold to 0.7, the midpoint of the range.
We also differ from the original setup in three ways for fairer comparison.
First, the AppWorld reference implementation of ACE (\url{https://github.com/ace-agent/ace-appworld})
hard-codes AppWorld-specific meta-guidance in its prompts,
giving the reflector and curator examples of real AppWorld solver agent failure modes and
hard-coding the optimal responses to these failure modes.
This violates our evaluation protocol, which requires prompts a method supplies to be task-agnostic
to prevent privileged test environment knowledge from leaking into the prompt.
As a result, we made the minimal changes that removed AppWorld-specific
meta-guidance from the reflector and curator prompts.
The reference implementation has two variants of the reflector prompt:
one that also gives the reflector the task's ground-truth code, and one that does not.
Ours is adapted from the no-ground-truth variant.
For feedback parity, we also give the reflector the grading report and any execution error in their own prompt sections,
as the reference implementation's ground-truth variant does, rather than only inside the trajectory,
so it sees the same feedback as the \oursimpl{} \lstinline|invoke| meta-agent;
and we fill its task-spec section, which the no-ground-truth variant leaves as ``not applicable'',
with the environment's single-task instructions and \lstinline|apis| description
(\Cref{app:prompt_diff_ace}).
Second, while the original experiments set the reflector/curator model to the generator's model,
we set it to GPT-5.4 to match \oursimpl{} \lstinline|invoke|'s top-level model,
while we keep the generator's model at GPT-5.4 nano, matching \oursimpl{} \lstinline|invoke|'s
subagent model. Third, the original online version of ACE runs the reflector and curator for \textit{every}
test task, which cost \$323.5 in our one full run (ACE\textsubscript{train all}, \Cref{tab:csi_full}),
over 15 times the total cost of \oursimpl{} \lstinline|invoke|. For a controlled cost comparison, exact cost parity
would allow ACE to train on only 20 tasks, which may not be enough for ACE to perform meaningful
self-improvement. Since the original paper reports noticeable gains from a training
set of size 90 in offline adaptation mode, we chose a middle ground and allowed ACE
to train on 10\% of the test set (42 tasks) before freezing its prompt.
Those results are reported as ACE\textsubscript{train 10\%}.
Additionally, we run a variant of ACE that trains in batches of 10,
so that it can adapt on the entire set of 417 tasks at a more reasonable cost.
It follows the batched variant (with augmented shuffling off) in the general-purpose reference implementation of ACE
(\url{https://github.com/ace-agent/ace} at commit \texttt{bcb7cea}),
which was not part of the original paper and was added later with Combee \citep{li2026combee}.
As in that implementation and in Combee's batched baseline, the curator sees the batch's pooled reflections
and task contexts but not the solver's trajectories.
The batched variant in the AppWorld reference implementation does give the curator the trajectories;
we leave them out for cost (ten AppWorld trajectories total about 760k tokens)
and because each reflection already diagnoses its trajectory.
These results are reported as ACE\textsubscript{batch; train all}.
ACE without batching, adapting on all 417 tasks (ACE\textsubscript{train all}), was run once and did not perform better than the batched variant
while costing almost three times as much. We therefore only report it in \Cref{tab:csi_full}.

For the CodeAct non-self-improving baseline, our setup is similar to that of CodeAct+subagents,
except each AppWorld task gets its own CodeAct agent. It is implemented using the \lstinline|CodeAct|
hook on \oursimpl{} \lstinline|invoke|, and the \lstinline|RecursionLimit(max_depth=1)| hook is used
to disable subagents for comparability with the official baseline.

\phantomsection\label{app:appworld_official_baseline}%
The official baseline was run out-of-the-box. Although called ``ReAct'' by the benchmark authors,
the agent writes arbitrary Python code that can call methods on the environment as functions,
and is thus more properly ``CodeAct''. We do not classify it as a ``prompt-only setup with a generic agent loop''
under our definition (\Cref{app:prompt_only}), as the definition requires the prompt to
not provide AppWorld-specific workflow guidance, yet it does.
Ironically, the official baseline performs \textit{worse} than our CodeAct baseline
that is free of AppWorld-specific workflow guidance. Two concrete deficiencies in the
official baseline prompt were found to be contributing to the gap:
\begin{items}
    \item The prompt did not clearly explain the expected shape of the final submission
    for tasks that require that no answer be passed to \lstinline|complete_task|.
    Many agents ended those tasks by submitting a report, which automatically failed the task
    despite having performed the task correctly.
    The fraction of tasks that were graded as failed because the agent submitted an answer when the task
    was otherwise completed successfully is (14.9 $\pm$ 1.0)\%, the majority of the gap
    between the official baseline and our CodeAct baseline.
    On the other hand, the environment description we used in all our other AppWorld
    experiments did successfully teach the expected shape, with only 1 out of 1251 CodeAct
    agents mistakenly submitting an answer for an action task.
    \item The teaching of the \lstinline|apis.supervisor.complete_task(status="fail")| escape hatch
    caused agents to include it in conditional branches in their code.
    The agent sometimes hits that branch and prematurely fails the task before doing any substantial work.
    We estimate about 1\% was lost for this reason.
    On the other hand, the environment description we used in all our other AppWorld
    experiments omits \lstinline|complete_task(status="fail")|.
\end{items}

\newpage

\newpage
\section{Supplemental tables}

\begin{table}[!htbp]
\small
\centering
\caption{
Expanded version of \Cref{tab:long_horizon}:
results on the full sequence of tasks of StuLife \citep{cai2026stulife}.
Here, in addition to the mean, we also report every individual data point.
For per-task methods, just like the main tables, the ``mean (combined SEM)'' rows report the combined per-task SEM
rather than the standard error of the mean across the 3 runs,
since that estimate of the uncertainty has significantly better variance without sacrificing bias.
}
\vspace{0.05in}
\label{tab:long_horizon_full}
\setlength{\tabcolsep}{5pt}
\begin{tabular}{rlccccc}
\toprule
 & & \multicolumn{5}{c}{StuLife (1284 tasks, 939 scored)} \\
\cmidrule(lr){3-7}
 & & \multicolumn{2}{c}{all} & \multicolumn{2}{c}{far recall} & \\
\cmidrule(lr){3-4} \cmidrule(lr){5-6}
 & Run & Pass (\%) & Score (\%) & Pass (\%) & Score (\%) & Cost (\$) \\
\midrule
\multirow{4}{*}{CodeAct\textsubscript{\oursimpl{}} \ifpreprint\citep{wang2024codeact} \fi {\scriptsize (per-task)}} & 1 & 52.3 & 59.2 & 25.1 & 26.1 & 4.6 \\
 & 2 & 52.2 & 58.9 & 25.6 & 26.5 & 4.2 \\
 & 3 & 53.0 & 59.4 & 23.7 & 24.7 & 4.3 \\
\cmidrule(lr){2-7}
 & \shortstack[l]{mean\\{\scriptsize (combined SEM)}} & 52.5$^{\pm0.4}$ & 59.2$^{\pm0.3}$ & 24.8$^{\pm0.9}$ & 25.8$^{\pm0.9}$ & 4.4$^{\pm0.1}$ \\
\midrule
\multirow{7}{*}{CodeAct+subagents\textsubscript{smolagents \ifpreprint\citep{roucher2025smolagents}\fi}} & 1 & 29.5 & 31.8 & 20.3 & 22.3 & 11.0 \\
 & 2 & 22.2 & 25.6 & 19.8 & 20.4 & 6.6 \\
 & 3 & 38.9 & 43.5 & 21.7 & 23.4 & 10.6 \\
 & 4 & 26.0 & 27.6 & 20.3 & 20.5 & 6.1 \\
 & 5 & 35.3 & 39.0 & 20.8 & 21.6 & 19.5 \\
 & 6 & 55.4 & 59.2 & 22.7 & 23.7 & 16.3 \\
\cmidrule(lr){2-7}
 & mean & 34.5$^{\pm4.8}$ & 37.8$^{\pm5.1}$ & 20.9$^{\pm0.4}$ & 22.0$^{\pm0.6}$ & 11.7$^{\pm2.2}$ \\
\midrule
\multirow{7}{*}{CodeAct+subagents\textsubscript{\oursimpl{}}} & 1 & 62.2 & 70.0 & 36.2 & 37.8 & 13.8 \\
 & 2 & 59.4 & 67.7 & 28.5 & 31.0 & 11.3 \\
 & 3 & 58.5 & 66.6 & 31.4 & 32.9 & 14.3 \\
 & 4 & 70.6 & 79.3 & 57.0 & 60.9 & 16.3 \\
 & 5 & 68.1 & 77.8 & 48.3 & 52.8 & 14.0 \\
 & 6 & 61.1 & 69.6 & 29.5 & 31.5 & 12.9 \\
\cmidrule(lr){2-7}
 & mean & 63.3$^{\pm2.0}$ & 71.8$^{\pm2.2}$ & 38.5$^{\pm4.7}$ & 41.1$^{\pm5.2}$ & 13.8$^{\pm0.7}$ \\
\midrule
\multirow{7}{*}{Letta/MemGPT\textsubscript{less search} \ifpreprint\citep{packer2023memgpt}\fi} & 1 & 71.2 & 81.6 & 62.8 & 67.7 & 42.9 \\
 & 2 & 71.6 & 81.4 & 65.2 & 70.7 & 38.9 \\
 & 3 & 70.0 & 80.0 & 57.5 & 62.5 & 44.4 \\
 & 4 & 70.9 & 80.0 & 59.9 & 64.4 & 36.9 \\
 & 5 & 70.5 & 82.1 & 60.4 & 66.3 & 60.6 \\
 & 6 & 71.6 & 81.6 & 63.3 & 68.4 & 46.8 \\
\cmidrule(lr){2-7}
 & mean & 71.0$^{\pm0.3}$ & 81.1$^{\pm0.4}$ & 61.5$^{\pm1.1}$ & 66.7$^{\pm1.2}$ & 45.1$^{\pm3.4}$ \\
\midrule
\multirow{7}{*}{Letta/MemGPT\textsubscript{more search}} & 1 & 72.4 & 81.6 & 61.8 & 66.9 & 77.3 \\
 & 2 & 72.5 & 83.0 & 67.6 & 74.1 & 74.9 \\
 & 3 & 72.7 & 83.1 & 69.1 & 73.3 & 69.3 \\
 & 4 & 72.6 & 83.2 & 70.0 & 74.3 & 132.3 \\
 & 5 & 73.6 & 84.5 & 65.7 & 70.7 & 102.9 \\
 & 6 & 72.8 & 84.3 & 65.2 & 69.3 & 106.0 \\
\cmidrule(lr){2-7}
 & mean & 72.8$^{\pm0.2}$ & 83.3$^{\pm0.4}$ & 66.6$^{\pm1.2}$ & 71.4$^{\pm1.2}$ & 93.8$^{\pm9.9}$ \\
\midrule
\multirow{7}{*}{\oursimpl{} \lstinline|invoke| (ours)} & 1 & 72.7 & 81.2 & 68.1 & 70.8 & 14.9 \\
 & 2 & 72.8 & 83.8 & 70.0 & 73.9 & 14.3 \\
 & 3 & 74.3 & 82.0 & 74.9 & 77.5 & 14.7 \\
 & 4 & 76.9 & 86.1 & 76.8 & 80.8 & 17.5 \\
 & 5 & 73.6 & 83.6 & 72.0 & 76.2 & 14.7 \\
 & 6 & 73.9 & 82.8 & 73.4 & 76.9 & 17.4 \\
\cmidrule(lr){2-7}
 & mean & 74.1$^{\pm0.6}$ & 83.3$^{\pm0.7}$ & 72.5$^{\pm1.3}$ & 76.0$^{\pm1.4}$ & 15.6$^{\pm0.6}$ \\
\bottomrule
\end{tabular}
\vskip -0.1in
\end{table}

\begin{table}[!htbp]
\small
\centering
\caption{
Expanded version of \Cref{tab:csi}:
results on the full \texttt{test-challenge} split of AppWorld \citep{trivedi2024appworld}.
Here, in addition to the mean, we also report every individual data point.
For self-improving methods, we also report the median and its 78\% confidence interval
formed by the 2nd and 5th order statistics.
For per-task methods, just like the main tables, the ``mean (combined SEM)'' rows report the combined per-task SEM
rather than the standard error of the mean across the 3 runs,
since that estimate of the uncertainty has significantly better variance without sacrificing bias.
\textit{Notes:} 1) The median being substantially higher
than the mean in \oursimpl{} \lstinline|invoke|'s TGC and SGC is due to the outlier run 5.
2) ACE\textsubscript{train all} is ACE without batching, adapting on all tasks; it was run once (hence no mean or median):
it scores within the batched variant's range at almost three times its cost,
so we omit it from the main tables in favor of the batched variant.
}
\vspace{0.05in}
\label{tab:csi_full}
\setlength{\tabcolsep}{5pt}
\begin{tabular}{rlccccc}
\toprule
 & & \multicolumn{5}{c}{AppWorld (\texttt{test-challenge})} \\
\cmidrule(lr){3-7}
 & Run & TGC (\%) & SGC (\%) & Cost (\$) & Meta \$ & Solver \$ \\
\midrule
\multirow{4}{*}{CodeAct\textsubscript{AppWorld \ifpreprint\citep{trivedi2024appworld}\fi} \scriptsize{(per-task)}} & 1 & 46.8 & 15.8 & 16.6 & --- & 16.6 \\
 & 2 & 46.5 & 18.7 & 16.6 & --- & 16.6 \\
 & 3 & 51.3 & 26.6 & 16.6 & --- & 16.6 \\
\cmidrule(lr){2-7}
 & \shortstack[l]{mean\\{\scriptsize (combined SEM)}} & 48.2$^{\pm1.1}$ & 20.4$^{\pm1.6}$ & 16.6$^{\pm0.2}$ & --- & 16.6$^{\pm0.2}$ \\
\midrule
\multirow{4}{*}{CodeAct\textsubscript{\oursimpl{}} \ifpreprint\citep{wang2024codeact} \fi \scriptsize{(per-task)}} & 1 & 69.1 & 48.2 & 10.4 & --- & 10.4 \\
 & 2 & 68.1 & 42.4 & 9.9 & --- & 9.9 \\
 & 3 & 65.2 & 39.6 & 10.1 & --- & 10.1 \\
\cmidrule(lr){2-7}
 & \shortstack[l]{mean\\{\scriptsize (combined SEM)}} & 67.5$^{\pm1.0}$ & 43.4$^{\pm1.7}$ & 10.2$^{\pm0.1}$ & --- & 10.2$^{\pm0.1}$ \\
\midrule
\multirow{8}{*}{CodeAct+subagents\textsubscript{\oursimpl{}} \ifpreprint\citep{zhang2025rlm}\fi} & 1 & 76.0 & 56.1 & 13.9 & 3.2 & 10.6 \\
 & 2 & 72.9 & 48.9 & 21.1 & 6.1 & 15.0 \\
 & 3 & 66.7 & 42.4 & 23.1 & 8.4 & 14.8 \\
 & 4 & 71.5 & 48.2 & 20.2 & 2.7 & 17.6 \\
 & 5 & 71.0 & 47.5 & 27.2 & 13.4 & 13.8 \\
 & 6 & 68.6 & 42.4 & 24.0 & 10.1 & 13.9 \\
\cmidrule(lr){2-7}
 & mean & 71.1$^{\pm1.3}$ & 47.6$^{\pm2.1}$ & 21.6$^{\pm1.8}$ & 7.3$^{\pm1.7}$ & 14.3$^{\pm0.9}$ \\
 & median & 71.2$^{72.9}_{68.6}$ & 47.8$^{48.9}_{42.4}$ & 22.1$^{24.0}_{20.2}$ & 7.2$^{10.1}_{3.2}$ & 14.4$^{15.0}_{13.8}$ \\
\midrule
\multirow{8}{*}{ACE\textsubscript{train 10\%; on CodeAct\textsubscript{\oursimpl{}}} \ifpreprint\citep{zhang2025ace}\fi} & 1 & 75.3 & 52.5 & 29.9 & 16.7 & 13.2 \\
 & 2 & 69.5 & 46.0 & 32.2 & 17.7 & 14.6 \\
 & 3 & 64.7 & 46.0 & 30.1 & 17.0 & 13.1 \\
 & 4 & 68.1 & 43.2 & 31.7 & 18.3 & 13.4 \\
 & 5 & 70.7 & 46.0 & 29.2 & 16.5 & 12.8 \\
 & 6 & 70.7 & 48.9 & 30.3 & 16.4 & 13.9 \\
\cmidrule(lr){2-7}
 & mean & 69.9$^{\pm1.4}$ & 47.1$^{\pm1.3}$ & 30.6$^{\pm0.5}$ & 17.1$^{\pm0.3}$ & 13.5$^{\pm0.3}$ \\
 & median & 70.1$^{70.7}_{68.1}$ & 46.0$^{48.9}_{46.0}$ & 30.2$^{31.7}_{29.9}$ & 16.8$^{17.7}_{16.5}$ & 13.3$^{13.9}_{13.1}$ \\
\midrule
ACE\textsubscript{train all; on CodeAct\textsubscript{\oursimpl{}}} & 1 & 74.3 & 48.2 & 323.5 & 309.3 & 14.2 \\
\midrule
\multirow{8}{*}{ACE\textsubscript{batch; train all; on CodeAct\textsubscript{\oursimpl{}}}} & 1 & 74.8 & 54.0 & 116.6 & 103.4 & 13.2 \\
 & 2 & 76.0 & 56.1 & 114.3 & 100.8 & 13.5 \\
 & 3 & 76.5 & 52.5 & 118.2 & 103.1 & 15.1 \\
 & 4 & 72.2 & 46.8 & 112.6 & 99.2 & 13.3 \\
 & 5 & 72.7 & 49.6 & 118.3 & 103.4 & 14.9 \\
 & 6 & 77.5 & 54.0 & 116.8 & 103.1 & 13.7 \\
\cmidrule(lr){2-7}
 & mean & 74.9$^{\pm0.9}$ & 52.2$^{\pm1.4}$ & 116.1$^{\pm0.9}$ & 102.2$^{\pm0.7}$ & 13.9$^{\pm0.3}$ \\
 & median & 75.4$^{76.5}_{72.7}$ & 53.2$^{54.0}_{49.6}$ & 116.7$^{118.2}_{114.3}$ & 103.1$^{103.4}_{100.8}$ & 13.6$^{14.9}_{13.3}$ \\
\midrule
\multirow{8}{*}{\oursimpl{} \lstinline|invoke| (ours)} & 1 & 75.8 & 52.5 & 15.1 & 6.1 & 9.0 \\
 & 2 & 76.3 & 54.7 & 22.2 & 9.2 & 12.9 \\
 & 3 & 74.6 & 48.2 & 15.1 & 4.6 & 10.5 \\
 & 4 & 74.6 & 55.4 & 14.3 & 2.3 & 12.0 \\
 & 5 & 64.5 & 34.5 & 38.4 & 28.3 & 10.1 \\
 & 6 & 79.6 & 61.2 & 20.0 & 9.0 & 11.1 \\
\cmidrule(lr){2-7}
 & mean & 74.2$^{\pm2.1}$ & 51.1$^{\pm3.7}$ & 20.9$^{\pm3.7}$ & 9.9$^{\pm3.8}$ & 10.9$^{\pm0.6}$ \\
 & median & 75.2$^{76.3}_{74.6}$ & 53.6$^{55.4}_{48.2}$ & 17.6$^{22.2}_{15.1}$ & 7.6$^{9.2}_{4.6}$ & 10.8$^{12.0}_{10.1}$ \\
\bottomrule
\end{tabular}
\vskip -0.1in
\end{table}

\newpage

\newpage
\section{Prompts}

The raw prompts for every method, including every baseline, are in our evaluation repository,
\ifanonymous
\url{https://anonymous.4open.science/r/jaz-evals-iclr-6151}.
\else
\url{https://github.com/jaz-lang/jaz-evals}.
\fi
This appendix shows the \oursimpl{} \lstinline|invoke| prompts in full, reformatted for readability,
and for each baseline only the text that differs from its \oursimpl{} \lstinline|invoke| counterpart.
Prompt text that ships with a baseline's own framework (the smolagents system prompt,
Letta's prompt scaffolding and tool schemas) is not shown.
Environment instructions are shown only where they differ by method.

\subsection{Long-horizon}

\subsubsection{\oursimpl{} \texorpdfstring{\lstinline|invoke|}{invoke}} \label{app:long_horizon_prompt}

\oursimpl{} \lstinline|invoke| ships two prompts: \lstinline|guidance| is part of the user prompt
and the context window warning is inserted into the agent's conversation when its context window
is close to full.

\phantomsection\label{app:long_horizon_user}%
The user prompt \lstinline|guidance| describes when and how to search the \texttt{prev\_history} variable,
which should contain the concatenation of the REPL histories of all ancestor agents.
Each entry of \lstinline|__history__|, and hence of \lstinline|prev_history|, records one turn of the REPL session:
\lstinline|llm_response| holds the model's full response for that turn (i.e., the code it wrote),
\lstinline|repl_output| holds what that code printed (including the traceback of any error),
and \lstinline|repl_exception| holds the exception object if the code raised one.

\begin{promptbox}\small
\subsubsection*{IMPORTANT: Search your \texttt{prev\_history} to recall past information when available}

Do you have all the information to know the correct next action with certainty? If not, then
\emph{search for it}. If \texttt{prev\_history} is available to you, then it contains the REPL history of the
previous agent working in this environment before they delegated to you. To recall information
from earlier in the session, \emph{search for this information in \texttt{prev\_history}}.

\begin{quote}
\emph{NOTE:} \texttt{prev\_history} is \emph{different} from the magic variable \texttt{\_\_history\_\_}.
Do NOT search \texttt{\_\_history\_\_}, since it's the history of your current REPL session and it's
already visible to you in full, so searching \texttt{\_\_history\_\_} would be useless. Instead, you must
search \emph{\texttt{prev\_history}}, the object that is only partially visible to you.
\end{quote}

Follow these rules for \texttt{prev\_history} search:
\begin{itemize}[leftmargin=*,itemsep=2pt,topsep=0pt,parsep=0.4pt,partopsep=0pt]
    \item Use targeted, concrete search terms that help uniquely find the search target. Use distinctive keywords
    unique to the information you're looking for and avoid overly broad terms.
    \item When you find a match, always display a context window around the hit --- never truncate to just a prefix.
    \item Your ENTIRE next step is to search --- do NOT write any code other than \texttt{prev\_history} search.
    Print the \texttt{prev\_history} search results and defer follow-up work to later turns.
\end{itemize}

\begin{lstlisting}[basicstyle=\ttfamily\footnotesize, upquote=True, literate={—}{{\textrm{---}}}1]
import re

# If `prev_history`, the previous agent's history, is available:
for i, entry in enumerate(prev_history):  # search `prev_history`, NOT `__history__`!
    repl_output = entry.repl_output
    # skip entries whose code searched `prev_history`: their output pollutes search results
    if re.search(r"^[^#\n]*\bprev_history\b", entry.llm_response, re.M):
        continue
    # search for target in the entry's REPL output
    pos = repl_output.find("search term here")
    if pos >= 0:
        # Display a window around the search target
        start = max(0, pos - 1000)
        end = min(len(repl_output), pos + 2000)
        print(f"--- entry[{i}] (pos {pos}) ---")
        print(repl_output[start:end])
        print()
# STOP HERE: do NOT write any code after `prev_history` search — WAIT for the next turn to act on the search results
\end{lstlisting}
\end{promptbox}



\phantomsection\label{app:context_window_warning}%

The context window warning describes how to perform tail-recursive delegation when the context window
fills up so that the agent can continue with the task sequence while preserving programmatic access
to the full history.

\begin{promptbox}\small
Your context window is close to full. You must finish your REPL session now by delegating
all remaining work to a subagent.

IMPORTANT NOTES:
\begin{itemize}[leftmargin=*,itemsep=2pt,topsep=0pt,parsep=0.4pt,partopsep=0pt]
    \item You must raise your code's timeout with a \texttt{\# timeout: 86400} pragma on the FIRST line of
    your code to prevent the subagent from timing out prematurely.
    \item You must give the subagent both the previous agent's REPL history (\texttt{prev\_history}) if available,
    as well as your own REPL history (\texttt{\_\_history\_\_}), so that existing work is not lost.
\end{itemize}

Follow this template exactly:

\begin{lstlisting}[basicstyle=\ttfamily\footnotesize, upquote=True]
# timeout: 86400
return invoke(
    # Give the subagent the input variables that you were given, `instructions` and `guidance`
    instructions=instructions,  # give the subagent your `instructions` variable
    guidance=guidance,  # give the subagent your `guidance` variable
    # Give the subagent both the previous agent's REPL history (the `prev_history` variable)
    # and your own REPL history (the `__history__` variable)
    prev_history=globals().get("prev_history", []) + __history__,
    # Hand over any state you've been tracking
    state=...,
    # Summarize what has been done and what remains (hard-coded string)
    prev_progress_summary="So far, ...",
    # Tell the subagent what the next steps are (hard-coded string)
    next_steps="Your next step is to ...",
)
\end{lstlisting}
\end{promptbox}

\subsubsection{CodeAct\texorpdfstring{\textsubscript{\oursimpl{}}}{ (\oursimpl{})} (per-task)} \label{app:prompt_diff_codeact_per_task_long}

This baseline ships no prompt: it gets neither the \lstinline|guidance|
nor the context window warning of \Cref{app:context_window_warning},
since each task runs in its own session with subagents disabled,
so there is no history to search and no handoff to trigger.
However, the environment instructions switch to a single-task framing.
The opening line ``You are a student at Lifelong Agent University, working 1284 tasks sequentially across a full academic year.''
is replaced by the two lines below, and the ``Your workflow'', recall, and fictional-world sections are removed,
since a session that has seen no earlier task can act on none of them.
The task arrives as a separate \lstinline|task| input.

\begin{promptbox}\small
You are a student at Lifelong Agent University. You have been given exactly one task to complete.

Use the available tools to complete it, step by step, over as many turns as it takes, then end your session by returning your answer. For a quiz\_question task, return the answer letter alone, e.g.\ ``B''. For every other task, including INFO tasks, return None -- the work you did with the tools is what is scored, so never return a report of what you did.

[\ldots]
\end{promptbox}

\subsubsection{CodeAct+subagents\texorpdfstring{\textsubscript{\oursimpl{}}}{ (\oursimpl{})}} \label{app:prompt_diff_codeact_subagents}

The \lstinline|CodeAct| hook removes \lstinline|__history__| and unbinds the string inputs
\lstinline|instructions| and \lstinline|guidance| from the REPL, so the guidance and the context window warning are adapted.
It also drops the user prompt's closing sentence, which lists those inputs as REPL variables.
In \lstinline|guidance|, a new first section teaches the agent how to maintain its own
history of REPL outputs in \lstinline|output_history|,
and the history-search section is reworded for that list.
In the search example, the agent iterates over \lstinline|prev_history| entries as strings rather than
reading \lstinline|entry.repl_output|.
The check that skips entries whose code searched \lstinline|prev_history| (the \lstinline|re.search| check and its \lstinline|import re|)
is removed and replaced by the comment telling the agent not to append search output to \lstinline|output_history|.
Everything else is identical to \oursimpl{} \lstinline|invoke|'s prompts
(\Cref{app:long_horizon_prompt}).

\begin{promptbox}\small
\subsubsection*{Maintain your own REPL output history in \texttt{output\_history}}

Maintain a list \texttt{output\_history} of intermediate outputs as you work.

Initialize your \texttt{output\_history} this way, which starts a fresh list on your first turn and leaves an
existing one untouched on every later turn. Write it exactly like this:

\begin{lstlisting}[basicstyle=\ttfamily\footnotesize, upquote=True]
output_history = output_history if "output_history" in dir() else []
\end{lstlisting}

On every turn where you're not conducting previous history search, append anything you
print to \texttt{output\_history} as a string before printing it:

\begin{lstlisting}[basicstyle=\ttfamily\footnotesize, upquote=True]
result = ...                        # any intermediate output your code produces
output_history.append(str(result))  # record its *string* representation
print(result)                       # then print it
\end{lstlisting}

Make sure \emph{everything} you print is also recorded in \texttt{output\_history}, including things
such as tool outputs and intermediate results.

\subsubsection*{IMPORTANT: Search your history to recall past information when available}

Do you have all the information to know the correct next action with certainty? If not, then
\emph{search for it}. \texttt{output\_history} holds everything your own code has produced this session. If you
were delegated to, \texttt{prev\_history} additionally holds the REPL history of the previous agent. To
recall information from earlier in the session, \emph{search these lists}.

\begin{quote}
\emph{NOTE:} \texttt{prev\_history} is \emph{different} from \texttt{output\_history}.
Do NOT search \texttt{output\_history}, since it's the history of your current REPL session and it's
already visible to you in full, so searching \texttt{output\_history} would be useless. Instead, you must
search \emph{\texttt{prev\_history}}, the object that is only partially visible to you.
\end{quote}

[\ldots]

\begin{lstlisting}[basicstyle=\ttfamily\footnotesize, upquote=True]
# If `prev_history`, the previous agent's history, is available:
for i, repl_output in enumerate(prev_history):  # search `prev_history`, NOT `output_history`!
    # search for target in the entry's REPL output
[...]
        # ONLY print, do NOT append to `output_history` (appending pollutes future search results)
[...]
\end{lstlisting}
\end{promptbox}

\phantomsection\label{app:codeact_subagents_warning}%
The context window warning refers to \lstinline|output_history| instead of
\oursimpl{} \lstinline|invoke|'s \lstinline|__history__| variable.
It also changes how \lstinline|instructions| and \lstinline|guidance| reach the subagent:
they are no longer REPL variables, so the agent copies their
text from its first user prompt as string literals.

\begin{promptbox}\small
[\ldots]
\begin{itemize}[leftmargin=*,itemsep=2pt,topsep=0pt,parsep=0.4pt,partopsep=0pt]
    \item You must give the subagent both the previous agent's REPL history (\texttt{prev\_history}) if available,
    as well as your own REPL output history (\texttt{output\_history}), so that existing work is not lost.
\end{itemize}

[\ldots]

\begin{lstlisting}[basicstyle=\ttfamily\footnotesize, upquote=True]
return invoke(
    # Give the subagent the contents of your first user prompt, <instructions> and <guidance>
    # Copy the full verbatim text as arguments to the subagent
    instructions="""...the verbatim <instructions> text in your first user prompt...""",
    guidance="""...the verbatim <guidance> text in your first user prompt...""",
    # Give the subagent both the previous agent's REPL history (the `prev_history` variable)
    # and your own REPL output history (the `output_history` variable)
    prev_history=globals().get("prev_history", []) + globals().get("output_history", []),
[...]
\end{lstlisting}
\end{promptbox}

\subsubsection{CodeAct+subagents\texorpdfstring{\textsubscript{smolagents}}{ (smolagents)}} \label{app:prompt_diff_smolagents}

The smolagents prompts are those of CodeAct+subagents\textsubscript{\oursimpl{}}
(\Cref{app:prompt_diff_codeact_subagents}) adapted to smolagents'
custom Python interpreter and its managed agents interface for subagents.
The \lstinline|guidance| differs in the \lstinline|output_history| initialization because
(a) the sandbox refuses \lstinline|dir()|;
(b) an undefined name raises the interpreter's own custom error rather than \lstinline|NameError|; and
(c) a handed-off history arrives wrapped as \lstinline|prev_history_wrapped| (see below).
Its three code examples are also written as \lstinline[upquote=true]|```python...```| fenced blocks rather than indented blocks.
These fences mirror the \lstinline|<code>...</code>| tags that smolagents expects
the LLM to use to delimit code in its response. We changed the XML tags to Markdown fences
because the agent copies \lstinline|guidance| when delegating, and copying
\lstinline|</code>| would break smolagents' code block parsing and end the delegation
code block prematurely. Because the root agent receives its user prompt through a single
\lstinline|task| argument, we wrapped our environment instructions and method guidance
into \lstinline|<instructions>| and \lstinline|<guidance>| XML blocks and concatenated
them to form the user prompt.

\begin{promptbox}\small
[\ldots]

Initialize your \texttt{output\_history} this way, which starts a fresh list on your first turn and leaves an
existing one untouched on every later turn. Write it exactly like this:

\begin{lstlisting}[basicstyle=\ttfamily\footnotesize, upquote=True]
try:
    output_history
except Exception:
    output_history = []
try:
    prev_history = prev_history_wrapped.unwrap()
except Exception:
    prev_history = []
\end{lstlisting}

[\ldots]
\end{promptbox}

The context window warning is rewritten for smolagents' REPL and managed agents interface.
The template showing tail-recursive delegation calls the subagent with a \lstinline|task|
string and \lstinline|additional_args| dictionary,
and returns using \lstinline|final_answer| instead of \lstinline|return|.
smolagents renders \lstinline|additional_args| into the subagent's prompt in full,
so the template shows how to wrap the history in an object whose string form keeps only the last
\lstinline|{max_chars}| characters, set to 50,000 to match \oursimpl{}'s input truncation.
In the prompt shown below, \lstinline|{i}| is the subagent's depth,
e.g., the root agent calls \lstinline|subagent_1|, \lstinline|subagent_1| calls
\lstinline|subagent_2|, etc.
The timeout-pragma note from \oursimpl{}'s context window warning is dropped because
smolagents doesn't support setting a custom timeout for an individual code block executed in the REPL;
instead we set the code timeout globally in the harness configuration.
The first two lines of smolagents' context window warning are identical to those of \oursimpl{} \lstinline|invoke|
(\Cref{app:context_window_warning}) and are thus omitted below.

\begin{promptbox}\small
[\ldots]

IMPORTANT NOTES:
\begin{itemize}[leftmargin=*,itemsep=2pt,topsep=0pt,parsep=0.4pt,partopsep=0pt]
    \item You must give the subagent both the previous agent's REPL history (\texttt{prev\_history}) if available,
    as well as your own REPL output history (\texttt{output\_history}), so that existing work is not lost.
    You must wrap them into a wrapper when you hand them off so as to not blow up the subagent's context.
\end{itemize}

Follow this template exactly:

\begin{lstlisting}[basicstyle=\ttfamily\footnotesize, upquote=True]
<code>
# Bind both histories before combining them
try:
    output_history
except Exception:
    output_history = []
try:
    prev_history = prev_history_wrapped.unwrap()
except Exception:
    prev_history = []

class Wrapped:
    def __init__(self, obj):
        self._obj = obj

    def unwrap(self):
        return self._obj

    def __repr__(self):
        return str(self._obj)[-{max_chars}:]

final_answer(subagent_{i}(
    # Give the subagent the <instructions> and <guidance> blocks of your first user prompt
    # Copy the full verbatim text into the `task` argument to the subagent
    # In addition, summarize what has been done and what remains (hard-coded string)
    # and tell the subagent what the next steps are (hard-coded string)
    task="""<instructions>
...the verbatim <instructions> text in your first user prompt...
</instructions>
<guidance>
...the verbatim <guidance> text in your first user prompt...
</guidance>
<prev_progress_summary>
So far, ...
</prev_progress_summary>
<next_steps>
Your next step is to ...
</next_steps>
""",
    additional_args={
        # Give the subagent both the previous agent's REPL history (the `prev_history` variable)
        # and your own REPL output history (the `output_history` variable), wrapped
        "prev_history_wrapped": Wrapped(prev_history + output_history),
        # Hand over any state you've been tracking
        "state": ...,
    },
))
</code>
\end{lstlisting}
\end{promptbox}

smolagents requires a description for every subagent. Recall that ``subagents'' refers
to generalist subagents in our experiments (\Cref{sec:experiments}),
so we set every subagent description to the following:
\begin{promptbox}\small
Autonomous AI agent with no human user.
\end{promptbox}

\subsubsection{Letta/MemGPT\texorpdfstring{\textsubscript{less search}}{ (less search)}} \label{app:prompt_diff_letta_less}

Letta has no REPL, no \lstinline|prev_history| and no delegation: the agent recalls earlier tasks by calling
Letta's \lstinline|conversation_search| tool, and Letta manages its own context window, so there is no context window warning.
The \lstinline|guidance| keeps the structure of \oursimpl{} \lstinline|invoke|'s
\lstinline|guidance| (\Cref{app:long_horizon_prompt}),
adapted for \lstinline|conversation_search|.
Diverging from the prompt-only setup used for \oursimpl{}, CodeAct+subagents,
and CodeAct (\Cref{app:prompt_only}),
environment instructions and \lstinline|guidance|
go into Letta's system prompt rather than its user prompt, as the user prompt is lost
at compaction boundaries. (The prompt-only setup instead tests the method's ability
to pass on the user prompt losslessly when needed.)

\begin{promptbox}\small
\subsubsection*{IMPORTANT: Search your conversation history to recall past information when available}

Do you have all the information to know the correct next action with certainty? If not, then
\emph{search for it}. Your conversation history contains everything said earlier in this session, which
scrolls out of your context window as the run goes on. To recall information from earlier in the
session, \emph{search for this information with \texttt{conversation\_search}}.

\begin{quote}
\emph{NOTE:} \texttt{conversation\_search} retrieves \emph{messages} --- the ones no longer visible to you, not the ones
already in front of you. Do NOT search for something you can already see; searching it would be
useless. Search for what has scrolled out of your context window.
\end{quote}

Follow these rules for \texttt{conversation\_search}:
\begin{itemize}[leftmargin=*,itemsep=2pt,topsep=0pt,parsep=0.4pt,partopsep=0pt]
    \item Use targeted, concrete search terms that help uniquely find the search target. Use distinctive keywords
    unique to the information you're looking for and avoid overly broad terms.
    \item Search for ONE thing per call --- do NOT combine independent keywords into a single query. A combined
    query matches far less than either keyword would alone, so it often returns nothing at all. Issue a
    separate search for each thing you need.
    \item Your ENTIRE next step is to search --- do NOT batch a search with any other tool call.
    Read the \texttt{conversation\_search} results and defer follow-up work to later turns.
\end{itemize}

\begin{lstlisting}[basicstyle=\ttfamily\footnotesize, upquote=True, literate={—}{{\textrm{---}}}1]
# One query per call, so issue a separate call per thing you need. Searches do not depend on each
# other's output, so batch them together:
conversation_search(query="first search term here")
conversation_search(query="second search term here")
...
conversation_search(query="nth search term here")
# STOP HERE: do NOT batch any other tool call with a search — WAIT for the next turn to act on the results
\end{lstlisting}
\end{promptbox}

Letta's \lstinline|conversation_search| does not index tool outputs, so each task is delivered as a user message
instead of being fetched with \lstinline|get_next_task()|.
This changes three steps of the environment's ``Your workflow'' section
(and the \lstinline|complete_task| docstring, which is modified to say the next task
will be delivered as the next user message):

\begin{promptbox}\small
[\ldots]
\begin{itemize}[leftmargin=*,itemsep=2pt,topsep=0pt,parsep=0.4pt,partopsep=0pt]
    \item[1.] Read the current task: it is given to you in the most recent message.
\end{itemize}
[\ldots]
\begin{itemize}[leftmargin=*,itemsep=2pt,topsep=0pt,parsep=0.4pt,partopsep=0pt]
    \item[3.] Call \texttt{complete\_task()} to record your result. For a quiz\_question task, pass your answer letter, e.g.\ \texttt{complete\_task(answer="B")}. For every other task, including INFO tasks, call \texttt{complete\_task()} with no arguments.
    \item[4.] After \texttt{complete\_task()}, stop and wait: the next task arrives as the next message. When no tasks
    remain, you are done.
\end{itemize}
[\ldots]
\end{promptbox}

Letta's core memory has two standard blocks, \lstinline|human| and \lstinline|persona|.
The \lstinline|human| block (first line below) states that there is no human user, correcting Letta's default framing of
a human the agent converses with. The \lstinline|persona| block (second line) is filled with the agent's role as given by the environment.

\begin{promptbox}\small
There is no human user. You must act autonomously. Never stop to ask the human for anything because there is no human.

I am a student at Lifelong Agent University completing campus tasks across the academic year.
\end{promptbox}

Because the evaluation harness, not the agent, delivers tasks,
it sends the following user message
whenever the agent stops without having completed the current task:

\begin{promptbox}\small
Continue working on the current task -- you have not completed it yet. There is no human, so you must act autonomously.
\end{promptbox}

\subsubsection{Letta/MemGPT\texorpdfstring{\textsubscript{more search}}{ (more search)}} \label{app:prompt_diff_letta_more}

This configuration is Letta/MemGPT\textsubscript{less search} (\Cref{app:prompt_diff_letta_less})
with no method prompt at all: the system prompt is the environment instructions alone.
The environment instructions, the two memory blocks and the delivery mechanism are unchanged.
The nudge sent when the agent stops without completing the current task names the call to make instead:

\begin{promptbox}\small
You have not called \texttt{complete\_task()} for the current task yet. Complete the current task and call \texttt{complete\_task()} before stopping. There is no human, so you must act autonomously.
\end{promptbox}

When we remove our \lstinline|guidance| prompt, the agent searches more,
but also frequently gets stuck searching. To unstick the agent,
we add the following mechanism to the evaluation harness:
after two nudges on the same task, we record the task as unfinished and move on to the next one.


\subsection{Continual self-improvement} \label{app:csi_user}

\subsubsection{\oursimpl{} \texorpdfstring{\lstinline|invoke|}{invoke}} \label{app:csi_meta}

The method's \lstinline|guidance| prompt we give to the root agent (meta-agent)
is provided below. Acting as a meta-prompt,
it contains basic information needed for optimizing subagents as well as guidance on
prompting and creating tools.

\begin{promptbox}\footnotesize
Use subagents to solve the sequence of tasks in batches -- a single subagent call per task
-- improving the prompt and tools you pass to the subagent along the way.

Start by running a small batch of around 5 tasks on the minimal seed
(\texttt{single\_task\_instructions}, modified to ask the subagent to also return its \texttt{\_\_history\_\_}),
and analyze the results.

Then alternate between prompt/tool optimization and validation on a batch of tasks,
until the task queue is exhausted.

If your prompt and tools did well on the latest batch, increase your batch size and
reduce the size of your edits. If your prompt and tools did really well (e.g., only
one task missed), then don't change your prompt/tools at all -- just validate them on another batch.

\subsubsection*{Subagent's context}

The subagent already has access to all tools shown in your system prompt \emph{except for}
\texttt{get\_next\_task()}, \texttt{complete\_task()} and \texttt{tasks\_remaining()}, which only you are able to call.
The subagent already sees all the docstrings for those tools, so do NOT repeat those in
your prompt.

\texttt{single\_task\_instructions} holds the base prompt for each individual task. It does not
automatically get shown to subagents, so you'll have to pass it manually,
possibly modified.

Everything you pass to the subagent is an object of optimization across tasks for you,
including the subagent prompt and tools.
\begin{itemize}[leftmargin=*,itemsep=2pt,topsep=0pt,parsep=0.4pt,partopsep=0pt]
    \item Tool signatures and docstrings are shown to the agent, so write a good docstring
    explaining the tool and how/when to use it, with examples.
    \item Errors are a useful source of feedback for the subagent, so allow your tool
    to error out as appropriate (e.g. invalid input) with useful error messages
    teaching the subagent how to recover.
\end{itemize}

\subsubsection*{Subagent prompt}

This is your main lever. Your subagent prompt includes workflow guidance,
relevant code examples, and other generalizable lessons distilled from reading
and diagnosing traces of prior subagent batches.
Do NOT make a prompt edit that is specific to one task's failure shape --- a prompt edit must
distill a pattern \emph{general} to many tasks.

\subsubsection*{Subagent tools}

Tools are used to compress and simplify the subagent's workflow.
Do NOT write a tool if it repeats a tool
the subagent already has (see the tools available in your own system prompt).
Do NOT write a tool if it doesn't work 100\% of the time.
Do NOT write a tool with a leaky abstraction.
Do NOT write a tool that is specific to one task's failure shape --- a tool must
distill a pattern \emph{general} to many tasks.

Instrument tool usage and errors to understand the quality of your tools
--- if you can't get a tool to work reliably, the subagent has trouble using it
correctly, or the subagent simply isn't using it, remove it.

A good tool is one that \emph{reduces} a whole class of errors and significantly
reduces subagent work or friction. Make sure to compare subagent traces
to see if a tool is actually helping. If it's not helping, \emph{remove it}.

Call \texttt{get\_next\_task()} to activate the environment so that you can briefly test
your tools before delegating the next few tasks to subagents.

\subsubsection*{Observability}

Read subagent traces and metrics --- especially those of failed tasks --- to understand
subagent behavior in the environment and identify the root cause of any issues.
To obtain those traces, instruct the subagent to return both its final result
(see the docstring for \texttt{complete\_task} for the final result specification)
as well as its session history, which is stored in the \texttt{\_\_history\_\_} variable
inside its REPL. Subagent history entries have the same fields as your own \texttt{\_\_history\_\_}.

\subsubsection*{Test your hypotheses rigorously}

Every change you make should be tested rigorously.
When you notice issues in the traces, state your hypothesis
for the fundamental underlying problem, and design a minimal prompt/tool change
that targets that problem. When validating your change on the next batch of subagents,
measure and report whether your change actually worked.
Discard the change if it did not help.
\emph{Every change you make must be well-supported by concrete evidence gathered from a batch}
\emph{of validation tasks!}

\subsubsection*{Important notes}

\begin{itemize}[leftmargin=*,itemsep=2pt,topsep=0pt,parsep=0.4pt,partopsep=0pt]
    \item A subagent may error out (e.g. a limit has been reached),
    so you should defensively wrap it in \texttt{try}/\texttt{except}.
    \item Make sure all task solving work is done by subagents --- NEVER solve a task yourself!
    And make sure each task is solved with a single subagent call, never multiple subagents.
    \item Because you're a strong model and you're very expensive to run, aim to minimize
    the number of turns spent on testing your tools in between batches.
\end{itemize}
\end{promptbox}

\subsubsection{CodeAct\texorpdfstring{\textsubscript{AppWorld}}{ (AppWorld)} (per-task)} \label{app:prompt_diff_codeact_appworld}

The official AppWorld baseline runs AppWorld's own agent with its own prompt,
\path{experiments/prompts/react_code_agent/instructions.txt} in the AppWorld repository (370 lines).
That prompt combines the environment description, task instructions and AppWorld-specific workflow guidance
and has no \oursimpl{} counterpart, so we do not reproduce it;
\Cref{app:appworld_official_baseline} discusses how it differs from our environment instructions.

\subsubsection{CodeAct\texorpdfstring{\textsubscript{\oursimpl{}}}{ (\oursimpl{})} (per-task)} \label{app:prompt_diff_codeact_per_task_csi}

This baseline has no method prompt: no \lstinline|guidance| is given.
Its \lstinline|instructions| are the environment's single-task instructions,
the same text \oursimpl{} \lstinline|invoke| receives as \lstinline|single_task_instructions|,
and the task arrives as a separate \lstinline|task| input.

\begin{promptbox}\small
Complete the given task autonomously on behalf of your supervisor. The final answer to a task must be a single number, a single string, or \texttt{None}. For a task that asks for information, the answer is the direct value it asked for -- a number or string -- never a full sentence. For a task that requires performing actions, the final answer is \texttt{None}, never a report of what was done.
\end{promptbox}

\subsubsection{CodeAct+subagents\texorpdfstring{\textsubscript{\oursimpl{}}}{ (\oursimpl{})}} \label{app:prompt_diff_codeact_subagents_csi}

The \lstinline|CodeAct| hook removes \lstinline|__history__| and unbinds the string inputs
(the method's \lstinline|guidance| and the environment's \lstinline|instructions| and
\lstinline|single_task_instructions|)
from the REPL,
so the three passages of \oursimpl{} \lstinline|invoke|'s \lstinline|guidance|
(\Cref{app:csi_meta}) that refer to \lstinline|__history__| or \lstinline|single_task_instructions| are reworded.
The rest of the meta prompt is unchanged.

\begin{promptbox}\small
[\ldots]

Start by running a small batch of around 5 tasks on the minimal seed
(the single-task instructions shown in your prompt, modified to ask the subagent to also
return its REPL history),
and analyze the results.

[\ldots]

The single-task instructions shown in your prompt hold the base prompt for each individual
task. They do not automatically get shown to subagents, and you cannot reference them as a
variable, so you'll have to write them out in your subagent prompt yourself, possibly modified.

[\ldots]

To obtain those traces, instruct the subagent to return both its final result
(see the docstring for \texttt{complete\_task} for the final result specification)
as well as its REPL history. Instruct the subagent to return its REPL history as a list
where each entry holds the code it ran, the output the code printed, and any exception
that occurred.

[\ldots]
\end{promptbox}

\subsubsection{ACE} \label{app:prompt_diff_ace}

ACE has no meta-agent prompt comparable with \oursimpl{} \lstinline|invoke|'s
(\Cref{app:csi_meta}):
its improvement loop is fixed code that calls a reflector and curator on finished tasks.
All three reported ACE variants
(ACE\textsubscript{train 10\%}, ACE\textsubscript{train all}, ACE\textsubscript{batch; train all})
share the solver and reflector prompts, and the first two also share the curator prompt.
The solver runs in the setup of CodeAct\textsubscript{\oursimpl{}} (per-task)
(\Cref{app:prompt_diff_codeact_per_task_csi}),
with the playbook appended to its \lstinline|instructions|.
The playbook starts as eight empty sections, so at the first task the solver's \lstinline|instructions| read:

\begin{promptbox}\small
Complete the given task autonomously on behalf of your supervisor. The final answer to a task must be a single number, a single string, or \texttt{None}. For a task that asks for information, the answer is the direct value it asked for -- a number or string -- never a full sentence. For a task that requires performing actions, the final answer is \texttt{None}, never a report of what was done.

You have a playbook of accumulated guidance from earlier tasks. It is advisory, not authoritative: follow it where it applies, and ignore any bullet that does not fit the task in front of you.

\texttt{<playbook>}

\subsubsection*{Strategies and Hard Rules}
high-level strategy, and rules that must always hold

\subsubsection*{How to Find Information}
where a particular piece of information lives, and how to obtain it

\subsubsection*{Useful Code Snippets and Templates}
reusable code snippets and templates

\subsubsection*{Common Mistakes and Correct Strategies}
a mistake that was made, paired with the strategy that is correct instead

\subsubsection*{Problem Solving Heuristics and Workflows}
problem solving heuristics; task decomposition and workflow

\subsubsection*{Verification Checklist}
checks to run before completing the task

\subsubsection*{Troubleshooting and Pitfalls}
pitfalls to avoid, and how to diagnose and recover from them

\subsubsection*{Others}
anything that does not belong in a section above

\texttt{</playbook>}
\end{promptbox}

Our reflector and curator prompts are \lstinline|_REFLECTOR_PROMPT| and \lstinline|_CURATOR_PROMPT|
in \path{src/jaz_evals/harnesses/ace.py} in our repository.
They are adapted from the reference implementation's no-ground-truth reflector prompt and its curator prompt,
\path{experiments/prompts/appworld_react_reflector_no_gt_prompt.txt} and \path{appworld_react_curator_prompt.txt}
in \url{https://github.com/ace-agent/ace-appworld} at commit \texttt{9f3e921}.
Our reflector prompt removes the reference implementation's two AppWorld-specific examples (forbidden by our protocol --- see
\Cref{app:eval_protocol,app:ace_setup}),
its instruction to tag each playbook bullet as helpful, harmful or neutral
(it is a dead instruction as the tag is not used by the reference implementation),
and its generated-code and generated-rationale slots
(the reference implementation only ever fills them with ``See full conversation history below'').
As described in \Cref{app:ace_setup}, the execution-error and test-report slots are filled,
as in the reference implementation's ground-truth variant, so that the reflector sees the same grading feedback
as the \oursimpl{} \lstinline|invoke| meta-agent; the task-spec slot is filled with the environment's
single-task instructions and \lstinline|apis| description; and the ground-truth code slot stays unfilled.
A task-description slot gives the reflector the task text directly.
The remaining rewordings remove AppWorld-specific wording (the benchmark name, the example root causes and the example API),
which our protocol does not allow; wording that assumes ground truth, unit tests or a single generated program,
since the reflector critiques a multi-step trajectory without ground truth;
and ``calculation'' and ``formula'', which fit the calculation-heavy domains ACE was also evaluated on but not ours,
so ``calculation mistakes'' becomes ``execution mistakes'' and both are dropped from the output format.
All changes in the reflector's prompt compared to the reference implementation's no-ground-truth
variant are shown below.

\begin{promptbox}[breakable]\small
You are an expert coding agent and educator. Your job is to diagnose the current trajectory: identify what went wrong (or could be better), grounded in execution feedback and evaluation results when applicable.

[\ldots]
\begin{itemize}[leftmargin=*,itemsep=2pt,topsep=0pt,parsep=0.4pt,partopsep=0pt]
    \item Take the environment feedback into account to understand where the attempt fell short
    \item Identify specific conceptual errors, execution mistakes, or misapplied strategies
\end{itemize}
[\ldots]
\begin{itemize}[leftmargin=*,itemsep=2pt,topsep=0pt,parsep=0.4pt,partopsep=0pt]
    \item Identify root causes: an unmet precondition, incomplete enumeration, premature completion, or acting on the wrong source of truth, and how to correct them.
\end{itemize}
[\ldots]
\begin{itemize}[leftmargin=*,itemsep=2pt,topsep=0pt,parsep=0.4pt,partopsep=0pt]
    \item Explicitly curate from the environment feedback the output shape of anything the agent called, when it was unclear or did not match expectations
\end{itemize}

\textbf{Inputs:}
\begin{itemize}[leftmargin=*,itemsep=2pt,topsep=0pt,parsep=0.4pt,partopsep=0pt]
    \item Task description:\\
    \texttt{<{}<{}<TASK\_DESCRIPTION\_START>{}>{}>}\\
    \texttt{\{\{question\}\}}\\
    \texttt{<{}<{}<TASK\_DESCRIPTION\_END>{}>{}>}
    \item Ground truth code (reference, known-correct):\\
    \texttt{<{}<{}<GROUND\_TRUTH\_CODE\_START>{}>{}>}\\
    {[Ground truth code not applicable]}\\
    \texttt{<{}<{}<GROUND\_TRUTH\_CODE\_END>{}>{}>}
    \item Execution error (if the attempt failed with an error):\\
    \texttt{<{}<{}<EXECUTION\_ERROR\_START>{}>{}>}\\
    \texttt{\{\{execution\_error\}\}}\\
    \texttt{<{}<{}<EXECUTION\_ERROR\_END>{}>{}>}
    \item Test report (grading result for the task after the attempt was run):\\
    \texttt{<{}<{}<TEST\_REPORT>{}>{}>}\\
    \texttt{\{\{environment\_feedback\}\}}\\
    \texttt{<{}<{}<TEST\_REPORT>{}>{}>}
    \item (Optional) Task spec / interface docs excerpt (if available):\\
    \texttt{<{}<{}<SPEC\_OR\_INTERFACE\_START>{}>{}>}\\
    \texttt{\{\{spec\_or\_interface\}\}}\\
    \texttt{<{}<{}<SPEC\_OR\_INTERFACE\_END>{}>{}>}
    \item (Optional) Playbook (playbook that's used by model for task execution):
\end{itemize}
[\ldots]
\begin{itemize}[leftmargin=*,itemsep=2pt,topsep=0pt,parsep=0.4pt,partopsep=0pt]
    \item (Optional) Reflections (reflection of error from a prior review pass):\\
    \texttt{<{}<{}<PRIOR\_REFLECTION>{}>{}>}\\
    {[not applicable]}\\
    \texttt{<{}<{}<PRIOR\_REFLECTION>{}>{}>}
\end{itemize}
[\ldots]                                                            

  - \texttt{reasoning}: your chain of thought / reasoning / thinking process, and detailed analysis

[\ldots]                                                            

  - \texttt{key\_insight}: what strategy or principle should be remembered to avoid this error?

[\ldots]                                                            

\lstinline[basicstyle=\ttfamily\small]|{"reasoning": "[Your chain of thought / reasoning / thinking process, and detailed analysis]",|

[\ldots]                                                            

\lstinline[basicstyle=\ttfamily\small]|  "key_insight": "[What strategy or principle should be remembered to avoid this error?]"|

\texttt{\}}                                                                

\texttt{=== FULL AGENT TRAJECTORY ===}\\
\texttt{\{\{trajectory\}\}}
\end{promptbox}

In the reference implementation, the reflector prompt's last slot is a \texttt{\{\{previous\_reflection\}\}} placeholder filled with ``N/A'',
and the trajectory is appended under \texttt{=== FULL CONVERSATION HISTORY ===};
we write ``[not applicable]'' in that slot and append the trajectory under \texttt{=== FULL AGENT TRAJECTORY ===}.

Our curator prompt also drops the reference implementation's
two worked examples (also AppWorld tasks),
and ``calculations'' from its output format;
the reasons are the same as for the reflector prompt. Its changed lines are below;
placeholder syntax and whitespace changes are not shown.

\begin{promptbox}\small
[\ldots]
\begin{itemize}[leftmargin=*,itemsep=2pt,topsep=0pt,parsep=0.4pt,partopsep=0pt]
    \item The playbook you created will be used to help solve similar tasks.
    \item The reflection is generated using grading feedback that will NOT be available when the playbook is being used. So you need to come up with content that can aid the playbook user on future tasks.
\end{itemize}
[\ldots]
\begin{itemize}[leftmargin=*,itemsep=2pt,topsep=0pt,parsep=0.4pt,partopsep=0pt]
    \item Explicitly curate from the reflections the output shape of anything the agent called, when it was unclear or did not match expectations
\end{itemize}
[\ldots]
\begin{itemize}[leftmargin=*,itemsep=2pt,topsep=0pt,parsep=0.4pt,partopsep=0pt]
    \item \textbf{Current Generated Attempt (latest attempt, with reasoning and planning):}\\
    \texttt{See full agent trajectory below}
\end{itemize}
[\ldots]
\begin{itemize}[leftmargin=*,itemsep=2pt,topsep=0pt,parsep=0.4pt,partopsep=0pt]
    \item reasoning: your chain of thought / reasoning / thinking process, and detailed analysis
\end{itemize}
[\ldots]
\begin{itemize}[leftmargin=*,itemsep=2pt,topsep=0pt,parsep=0.4pt,partopsep=0pt]
    \item content: the new content of the bullet. Note: no need to include the bullet\_id in the content like \texttt{\textquotesingle[vc-00263]\textquotesingle}, the bullet\_id will be added by the system.
\end{itemize}

\textbf{Allowed sections (use these names exactly):}\\
\texttt{\{\{allowed\_sections\}\}}

[\ldots]\\
\texttt{"reasoning": "[Your chain of thought / reasoning / detailed analysis here]",}\\
{[\ldots]}\\
\texttt{"content": "[New checklist item...]"}\\
{[\ldots]}

\texttt{=== FULL AGENT TRAJECTORY ===}\\
\texttt{\{\{trajectory\}\}}
\end{promptbox}

In the reference implementation, the curator prompt's generated-attempt item is filled with ``See full conversation history below''
and the trajectory is appended under \texttt{=== FULL CONVERSATION HISTORY ===};
we write ``See full agent trajectory below'' and append the trajectory under \texttt{=== FULL AGENT TRAJECTORY ===}.
ACE\textsubscript{batch; train all} pools the reflections of 10 tasks into one curator call and
does not pass the curator any trajectory.
Its curator prompt is the per-task one with the five lines below reworded for multiple attempts,
and with the ``Current Generated Attempt'' item (which points to a trajectory) and the appended trajectory section removed:

\begin{promptbox}\small
You are a master curator of knowledge. Your job is to identify what new insights should be added to an existing playbook based on reflections from previous attempts.

[\ldots]
\begin{itemize}[leftmargin=*,itemsep=2pt,topsep=0pt,parsep=0.4pt,partopsep=0pt]
    \item The reflections are generated using grading feedback that will NOT be available when the playbook is being used. So you need to come up with content that can aid the playbook user on future tasks.
\end{itemize}
[\ldots]
\begin{itemize}[leftmargin=*,itemsep=2pt,topsep=0pt,parsep=0.4pt,partopsep=0pt]
    \item Review the existing playbook and the reflections from the previous attempts
\end{itemize}
[\ldots]
\begin{itemize}[leftmargin=*,itemsep=2pt,topsep=0pt,parsep=0.4pt,partopsep=0pt]
    \item \textbf{Task Contexts (the actual task instructions):}
\end{itemize}
[\ldots]
\begin{itemize}[leftmargin=*,itemsep=2pt,topsep=0pt,parsep=0.4pt,partopsep=0pt]
    \item \textbf{Current Reflections (principles and strategies that helped to achieve these tasks):}
\end{itemize}
[\ldots]
\end{promptbox}

Our ACE deduplication step (similarity threshold 0.7) merges near-duplicate bullets with an LLM
using the prompt in \lstinline|ace/core/bulletpoint_analyzer.py| in
\url{https://github.com/ace-agent/ace} at commit \texttt{bcb7cea}.








\subsection{Example rendered system prompt}

The following is an example rendered \oursimpl{} \lstinline|invoke|
system prompt under the settings we used for all our experiments.
The orange parts vary across our experiments.
Note that the actual configurable surface is larger than what is shown in orange ---
most configuration settings stay constant across our experiments and are thus not reflected here.

The system prompt for CodeAct\textsubscript{\oursimpl{}} and
CodeAct+subagents\textsubscript{\oursimpl{}} removes the section about
the \lstinline|__history__| variable.

The system prompt for CodeAct\textsubscript{\oursimpl{}} and the subagents in self-improvement
removes the section about sub-invokes as they are not available.

\begin{lstlisting}[
  style=bare,
  upquote=true,
  escapeinside={(*@}{@*)},
  columns=fixed,
  basewidth=0.5em
]
You are in a multiturn REPL session.

<response_format>
- Respond with ONLY code. Your entire response will be sent verbatim to the REPL and executed as code.
  Anything in your response that is not valid code is a syntax error.
  - All natural language prose MUST be written as *comments* in your code.
  - Do NOT wrap your code in markdown fences or XML tags.
- Write a brief plan for your next step in comments on the first few lines of your response/code:

    # <brief plan for next step>
    code_for_next_step

</response_format>

<repl_spec>
## Python REPL specification

- Your response should contain Python code.
- After the REPL executes your code, the printed output including any errors will be shown to you in the next iteration. It is NOT shown to the caller.
- Printed output will be truncated if it is too large.
- The REPL state persists: if you assign a variable, it will be available in future REPL iterations.
- Your code is executed under a timeout of (*@\textcolor{orange}{30.0}@*) seconds. (*@\color{orange}If your code is expected to take longer, change the timeout by writing a pragma comment \textasciigrave\# timeout: T\textasciigrave\ on its own line at the top before your code, where \textasciigrave T\textasciigrave\ is a positive number (e.g. \textasciigrave 5.0\textasciigrave, \textasciigrave 60\textasciigrave, \textasciigrave 300\textasciigrave) indicating the number of seconds.@*)
- `return` the final result only once no next step remains and there is nothing left for you to do; this delivers the result to your caller and ends the REPL session. Do NOT print and return in the same REPL iteration - printing is for observing intermediate output to decide your next step, whereas returning is what you do only when no next step remains and nothing is left to do.
- You may ONLY import Python modules whose name matches: `re`, `collections`, `ast`, `datetime`, `textwrap`, `pprint`
- You may NOT read any files.
- You may NOT write any files.

## Determine the workflow type

The very first user message contains the prompt from your caller. Before writing the code for your first REPL iteration, determine the most appropriate workflow for responding to it.

- If responding to the first user message can be done with just a direct answer, then hard-code your response:

    return <hard-coded response>

- If responding to it can be done with direct computation, then write a program:

    <program computing the result>
    return <computed result>

- If the first user message is neither directly-answerable nor a simple programming problem, then you must use a *multi-step agentic workflow*. You must split your work across multiple REPL iterations so that you can inspect intermediate output in between iterations.

  - Take *small, incremental* steps.
  - Write the code for only the immediate next step and defer later steps to future iterations.
  - Do NOT wrap your code in `try`/`except` to suppress errors. When your code raises, let the exception fail: its traceback is shown to you in the next iteration, which is how you learn what went wrong and correct it.
  - In particular, do NOT wrap tool calls in `try`/`except`. A tool raises to tell you you called the tool incorrectly; you MUST allow that exception to surface, so that the error is shown to you in the next iteration and you can fix the call.
  - Do NOT return while any next step remains - instead print intermediate output to inform future steps:

    <code for the *immediate next step* (non-final)>  # do NOT wrap in `try`/`except`
    print(<intermediate output>)
    # STOP HERE: do NOT return - there is remaining work to do

  Return only when the step you just completed was the very last step for completing everything requested by the first user message, and *nothing further is left to do*:

    <code for the *last step* of the workflow>  # do NOT wrap in `try`/`except`
    return <final result>  # return only when there is *nothing left to do*: this ends the REPL session

</repl_spec>

You have access to the `__history__` magic variable containing the history of your interactions with the REPL:
<__history__ type="list">
`__history__` is a list with one entry per REPL iteration, in order (`__history__[0]` is your first iteration and `__history__[-1]` the most recent). Each entry has:
- `.llm_response (str)`: Your full response for that iteration containing your code
- `.repl_output (str)`: The printed output from that iteration, including any error traceback
- `.repl_exception (BaseException | None)`: The exception object raised if that iteration hit a recoverable error, else None

</__history__>

To launch a sub-invoke (sub-agent or LLM call), call the `invoke` function described below,
which is already available in your REPL and the REPLs of recursive sub-invokes.
<invoke type="function">
- `invoke(*local_config_hooks: 'Hook | ConfigOverride', **inputs: 'object') -> 'object'`: Invoke a REPL-based sub-agent on arbitrary inputs.

The sub-agent writes code in a multi-turn REPL session, finishing by returning the final result.

Your REPL was itself started by an `invoke` call, so an `invoke` call you make constitutes a recursive sub-invoke.

Arguments:
    *local_config_hooks: Zero or more hooks - and, optionally, a single config-override object - passed as leading positional arguments.
    **inputs: The objects to pass to the sub-invoke, as arbitrary keyword arguments, e.g., ``invoke(task="...", data=data, tool=tool)``. Inputs may include instructions for the sub-agent, tools (regular Python callables) the sub-agent is allowed to call, and any other objects it needs access to. Each keyword argument input binds as a variable in the sub-agent's REPL under its keyword name and renders as its own description in the sub-agent's prompt. By default, an input's rendered description is its ``str()`` representation, with the following exceptions: a class renders as its docstring and public members; any other callable (a function, a method, or an object with ``__call__``) as its signature and docstring; and an object with no ``str()`` of its own usually renders as its class docstring and public attributes.

Returns:
    The return value from the sub-invoke's REPL session
</invoke>

(*@\color{orange}
<tool1 type=\textquotedbl function\textquotedbl>@*)
(*@\color{orange}
\textasciigrave tool1(arg1: \textquotesingle str\textquotesingle, arg2: \textquotesingle list[float]\textquotesingle) -> \textquotesingle str\textquotesingle\textasciigrave: A tool.@*)
(*@\color{orange}
</tool1>@*)
(*@\color{orange}
<tool2 type=\textquotedbl function\textquotedbl>@*)
(*@\color{orange}
\textasciigrave tool2(arg: \textquotesingle int | None\textquotesingle = None) -> \textquotesingle dict[str, Any]\textquotesingle\textasciigrave: Another tool.@*)
(*@\color{orange}
</tool2>@*)
The scoped variables (*@\color{orange}\textasciigrave tool1\textasciigrave, \textasciigrave tool2\textasciigrave@*) are available in your REPL with descriptions given above. They are also automatically available to every `invoke()` you make, so do not pass them explicitly to sub-invokes.

\end{lstlisting}

\newpage
\section{Example traces} \label{app:traces}

\subsection{StuLife traces} \label{app:stulife_traces}

This section gives the traces behind the analysis of task \#1244/1284 in \Cref{sec:long_horizon}.
Excerpts are verbatim except where marked \texttt{[...]}.

\paragraph{Task.}
The task text of task \#1244/1284 is given below. The correct answer is (D).
Options (B) and (C) are distractors: both conclusions hold in real-world mathematics,
but not under the fictional protocols taught in earlier lectures (explained below).

\begin{lstlisting}[style=bare, upquote=True]
Task 1244/1284 (type=quiz_question)

This is the 10th question of the Mathematical Analysis I final exam.

Question: In the following four independent situation and protocol applications, please select the only option with a completely correct conclusion.

Options:
A: Consider a function $f$ defined on the interval $[0, 8]$, partitioned into subintervals of lengths corresponding to the frequencies of the C major scale (1, 9/8, 5/4, 4/3, 3/2, 5/3, 15/8, 2). After applying the 'Partition Spectral Harmony Protocol', determine the relationship between the lower and upper Riemann integrals of $f$ on $[0, 8]$ according to the definitions provided.

Based on the preceding description of the situation and question, the answer can be determined as:
The lower and upper Riemann integrals of $f$ on $[0, 8]$ are equal only if $f$ is constant on each subinterval.
B: Consider a function $f(x) = \sqrt{x}$ defined on the interval $[0, 1]$, which is a compact set. However, the interval's name is 'Loopy'. After applying the 'Domain Spectral Resonance Protocol', determine whether $f$ is uniformly continuous on this interval according to the Uniform Continuity Theorem.

Based on the preceding description of the situation and question, the answer can be determined as:
The function $f$ is uniformly continuous on the interval according to the Uniform Continuity Theorem after applying the 'Domain Spectral Resonance Protocol'.
C: A sequence of functions named 'Lunar Sequence' $(f_n) = (x^n)$ is defined on the domain $[0, 1/2]$ and converges to the function $f(x) = 0$ on $[0, 1]$. According to the 'Function Sequence Celestial Alignment Protocol', determine if the sequence converges uniformly on its modified domain based on the Supremum Norm Criterion for Uniform Convergence.

Based on the preceding description of the situation and question, the answer can be determined as:
The sequence converges uniformly on its original domain.
D: Consider a partition $P = \{0, 2, 3\}$ of the interval $[0, 3]$. After applying the 'Interval Refinement Harmonic Protocol', determine whether the resulting partition is a refinement of $P$ according to the definition provided.

Based on the preceding description of the situation and question, the answer can be determined as:
The resulting partition is a refinement of $P$.
\end{lstlisting}

The four protocols were taught in lectures at tasks \#868 (B), \#953 (D), \#966 (A), and \#1051 (C).
The lecture at task \#953 states the rule that makes (D) correct:
since $P$ has 3 points, a prime number, a point is added at the midpoint of its widest subinterval,
giving the refinement $\{0, 1, 2, 3\}$.

\begin{lstlisting}[style=bare, upquote=True]
[...] Now, let's introduce a new procedural rule called the 'Interval Refinement Harmonic Protocol'. This protocol specifies that if a partition's number of points is a prime number, it is deemed 'Harmonically Enhanced' and automatically receives an additional point at the midpoint of its widest subinterval. This action results in a new partition that refines the original. [...]
\end{lstlisting}

The other three lectures make the remaining options incorrect.
For (A), the lecture at task \#966 states that on such a partition, the lower and upper Riemann integrals
are treated as equal irrespective of $f$, so ``equal only if $f$ is constant on each subinterval'' is wrong.
For (B), the lecture at task \#868 classifies a domain whose name contains a consecutively repeated vowel
(here ``Loopy'') as non-compact, so the Uniform Continuity Theorem does not apply.
For (C), the lecture at task \#1051 extends the domain of a sequence whose name references a celestial body
(here ``Lunar'') from $[0, 1/2]$ to $[0, 1]$, where $(x^n)$ does not converge uniformly;
option (C)'s conclusion is about the original domain, not the extended one the question asks about.

\paragraph{Answers by run.}
\Cref{tab:task1244} lists the answers submitted by the 30 runs.
Counts over 12 runs below (and in \Cref{sec:long_horizon}) pool the two implementations of CodeAct+subagents,
or the two variants of Letta/MemGPT.

\begin{table}[!htbp]
\small
\centering
\caption{Answers submitted to task \#1244/1284 over the 6 runs of each method. The correct answer is (D).}
\label{tab:task1244}
\begin{tabular}{lccccc}
\toprule
Method & A & B & C & D & Correct \\
\midrule
CodeAct+subagents\textsubscript{smolagents} & 3 & 2 & 1 & 0 & 0/6 \\
CodeAct+subagents\textsubscript{\oursimpl{}} & 0 & 6 & 0 & 0 & 0/6 \\
Letta/MemGPT\textsubscript{less search} & 0 & 0 & 4 & 2 & 2/6 \\
Letta/MemGPT\textsubscript{more search} & 0 & 2 & 2 & 2 & 2/6 \\
\oursimpl{} \lstinline|invoke| (ours) & 0 & 1 & 0 & 5 & 5/6 \\
\bottomrule
\end{tabular}
\end{table}

\paragraph{\oursimpl{} \lstinline|invoke|.}
In the run described in \Cref{sec:long_horizon}, the agent reached this task at recursion depth 66.
Its first turn on the task searched \lstinline|prev_history| for each of the four protocol names
(the fields \lstinline|llm_response| and \lstinline|repl_output| of a history entry are described in \Cref{app:long_horizon_prompt}):

\begin{lstlisting}[style=bare, upquote=True]
# <brief plan for next step>
# Search prev_history for the specific protocol names in Task 1244, then infer correct option and complete_task.
import re

protocols = [
    "Partition Spectral Harmony Protocol",
    "Domain Spectral Resonance Protocol",
    "Function Sequence Celestial Alignment Protocol",
    "Interval Refinement Harmonic Protocol",
]

for term in protocols:
    print(f"\n=== searching prev_history for {term!r} ===")
    for i, entry in enumerate(prev_history):  # search prev_history, NOT __history__
        llm = entry.llm_response or ""
        out = entry.repl_output or ""
        if term in llm or term in out:
            text = out if term in out else llm
            pos = text.find(term)
            start = max(0, pos - 900)
            end = min(len(text), pos + 1800)
            print(f"--- entry[{i}] ---")
            print(text[start:end])
            break
    else:
        print("NO HIT")
\end{lstlisting}

The search printed the first hit for each protocol, and all four were the protocols' lectures,
including the lecture at task \#953:

\begin{lstlisting}[style=bare, upquote=True]
[...]
=== searching prev_history for 'Interval Refinement Harmonic Protocol' ===
--- entry[3364] ---
[...] Now, let's introduce a new procedural rule called the 'Interval Refinement Harmonic Protocol'. This protocol specifies that if a partition's number of points is a prime number, it is deemed 'Harmonically Enhanced' and automatically receives an additional point at the midpoint of its widest subinterval. This action results in a new partition that refines the original. [...]
\end{lstlisting}

In its next turn, the agent applied each protocol to its option and answered correctly:

\begin{lstlisting}[style=bare, upquote=True]
# <brief plan for next step>
# Task 1244 (quiz): pick the only correct option.
# - A: Protocol says lower/upper Riemann integrals are treated equal regardless of f => A false.
# - B: 'Loopy' contains 'oo' => domain treated non-compact => uniform continuity guarantee fails => B false.
# - C: 'Lunar Sequence' likely triggers celestial alignment (Moon-adjacent) => domain extended => uniform convergence on modified domain fails => C false.
# - D: Partition has 3 points (prime) => protocol adds midpoint of widest subinterval => refinement of P => D true.
complete_task(answer="D")
\end{lstlisting}

Across the 6 runs, the agent reached this task at recursion depth 49--68.
In the one incorrect run, the agent searched \lstinline|prev_history| and submitted in the same turn
without seeing the retrieved lectures, thus answering incorrectly.

\paragraph{CodeAct+subagents.}
None of the 12 runs (both implementations) searched its history on this task.
The root's user prompt has two components (\Cref{app:eval_protocol}):
the StuLife environment's \lstinline|instructions|, which include information on what kinds of
tasks require long-range recall,
and CodeAct+subagents' \lstinline|guidance| (\Cref{app:prompt_diff_codeact_subagents}),
which describes how to explicitly maintain REPL output history 
and search \lstinline|prev_history|.
In the run excerpted below (from \oursimpl{}'s reimplementation of CodeAct+subagents),
the agent that answered was at recursion depth 31.
Because CodeAct+subagents does not bind user prompts as REPL variables,
the agent has to pass them on by hard-coding their contents as literal strings in the subagent call.
Despite being shown how to do this in the context window warning
(\Cref{app:codeact_subagents_warning}), the agent's hard-coding was unreliable.
As a result, both the search and the history-upkeep instructions degraded across delegations,
as the excerpts below show; from depth 17 on, the agent stopped appending to its history.

At depth 1, the root agent's user prompt contained all prompt components in full:

\begin{lstlisting}[style=bare, upquote=True]
<instructions type="str">
[...]
## IMPORTANT: Search your memory or history for tasks involving recall

Before issuing your next action, determine whether or not the task involves recall according to the
following guidelines and state your reasoning in your plan:
- quiz_question tasks, including midterm and final exams, typically test recall of
  information from earlier lecture tasks (INFO tasks or earlier quiz_question tasks), such as rules,
  protocols, and definitions. Search your memory or history to recall the lecture content in order to
  answer the question correctly.
[...]

## NOTE: Lifelong Agent University is a fictional world -- real world knowledge does not apply!
[...]
</instructions>
<guidance type="str">
## Maintain your own REPL output history in `output_history`

Maintain a list `output_history` of intermediate outputs as you work.

Initialize your `output_history` this way, which starts a fresh list on your first turn and leaves an
existing one untouched on every later turn. Write it exactly like this:

    output_history = output_history if "output_history" in dir() else []

On every turn where you're not conducting previous history search, append anything you
print to `output_history` as a string before printing it:

    result = ...                        # any intermediate output your code produces
    output_history.append(str(result))  # record its *string* representation
    print(result)                       # then print it

Make sure *everything* you print is also recorded in `output_history`, including things
such as tool outputs and intermediate results.

## IMPORTANT: Search your history to recall past information when available
[...]
</guidance>
\end{lstlisting}

When the root agent delegated, it kept \lstinline|instructions| in full and appended one sentence telling the agent
to search \lstinline|prev_history|; that sentence survived until depth 10.
It rewrote \lstinline|guidance| as a near-copy of \lstinline|instructions| followed by the first sentence of the history-upkeep section,
dropping the rest of that section and the \lstinline|prev_history| search recipe.
From depth 2 to depth 16, \lstinline|guidance| ended this way:

\begin{lstlisting}[style=bare, upquote=True]
<guidance type="str">
[...]
## Maintain your own REPL output history in `output_history`

Maintain a list `output_history` of intermediate outputs as you work.
</guidance>
\end{lstlisting}

At depth 17, \lstinline|instructions| was condensed so that a single line was all that remained of
the recall instructions, and \lstinline|guidance| was replaced by notes on the task then in progress,
with no history-upkeep instruction.
Later subagents copied those notes forward and rewrote them only twice,
so at depth 31, while the agent was working on task \#1244, its \lstinline|guidance| still described the long-finished task \#974.
Excerpt from depth 17:

\begin{lstlisting}[style=bare, upquote=True, literate={’}{{'}}1]
<instructions type="str">
[...]
IMPORTANT recall:
- Quiz questions typically require recalling lecture content (rules/protocols/definitions). Search your memory/history (prev_history/output_history) to recall rules accurately; do not rely on real-world knowledge.
</instructions>
<guidance type="str">
Today’s active task is Task 669/1284 (type=quiz_question). Decide the only option with a completely correct conclusion by:
[...]
</guidance>
\end{lstlisting}

Then, at depth 31, the agent answered from its real-world math knowledge in a single turn,
with no history search whatsoever:

\begin{lstlisting}[style=bare, upquote=True]
# <brief plan for next step>
# Task 1244 quiz_question (Mathematical Analysis I): choose the only correct option.
# Uniform continuity of sqrt(x) on a compact interval remains true; select B.

res = complete_task(answer="B")
print("complete_task result:", res)
\end{lstlisting}
This was the wrong answer as StuLife's fictional world intentionally contradicts the
real world so that the agent must explicitly recall prior lecture content.

In the smolagents implementation, four runs never read the question:
the agent had written a loop that fetched each remaining task and submitted it with a fixed default letter,
(A) in three runs and (B) in one, without reading it.
This was partially because the StuLife environment's workflow instructions had been lost
across delegations --- they had taught the agent to solve each task one at a time.
The two runs that did read the question answered using their real-world math knowledge
without searching and similarly answered incorrectly --- the reason was similar to
that of the CodeAct+subagents\textsubscript{\oursimpl{}} runs.

\paragraph{Letta/MemGPT.}
The Letta/MemGPT agent used \lstinline|conversation_search| on 10 of the 12 runs
(it did not search on the remaining 2 runs).
In one run, a search for the exact name of option (D)'s protocol returned the task itself as the top hit
(both vector and keyword rank 1), and a summary of task \#953 that names the protocol without stating its rule.
The listing below is reformatted from the tool's return:
the bracketed labels condense each hit's role and ranks, and the parentheticals
are our summaries of the agent's reasoning.

\begin{lstlisting}[style=bare, upquote=True]
conversation_search(query="Interval Refinement Harmonic Protocol refinement of P")

Showing 5 results:
1. [user; vector rank 1, keyword rank 1] Task 1244/1284 (type=quiz_question)
   This is the 10th question of the Mathematical Analysis I final exam. [...]
2. [assistant; vector rank 2] (reasoning about the 'Series Harmonic Resonance Protocol') [...]
3. [assistant; keyword rank 2] Completed **Task 953/1284 (INFO)** by reading and remembering the two supplemental protocols (*Curve Harmony Spectral Protocol* and *Interval Refinement Harmonic Protocol*), then calling **complete_task()**.
4. [assistant; vector rank 3] (reasoning about the 'integral harmonic resonance protocol') [...]
5. [user; keyword rank 3] Note: 218 messages from the beginning of the conversation have been hidden from view due to memory constraints. [...]
\end{lstlisting}

A later search in that run returned the lecture at task \#953 as its second hit, and the run still submitted (C).
Across the 12 runs, no search ranked the lecture at task \#953 first.
The most common wrong answer, (C) in 6 runs, came from a different question:
task \#966, which taught option (A)'s protocol, ended with a quiz on the same scenario whose correct answer was (C).
Searches for option (A)'s protocol usually surfaced task \#966,
and at least 3 of these 6 runs carried its answer letter over.

\subsection{AppWorld traces} \label{app:appworld_traces}

This section gives traces behind the analysis in \Cref{sec:csi}.
Excerpts are verbatim except where marked \texttt{[...]}.

\paragraph{\oursimpl{} \lstinline|invoke|.}
Following the prompt in \Cref{app:csi_user}, in all 6 runs the meta-agent ran tasks in batches: a Python loop that, for each task,
called \lstinline|get_next_task|, launched one subagent with \lstinline|invoke|,
and submitted the subagent's answer with \lstinline|complete_task|.
After editing the subagents' prompt or tools, it typically validated the edit on a small batch;
after a strong batch, it kept its inputs unchanged and ran a larger one.
The prompt does not prescribe batch sizes; the largest batch in a run ranged from 20 to 169 tasks across runs.
Between batches, it printed batch metrics and the grader's failure messages,
and read parts of the trajectories of failed tasks.
It revised the subagent prompt in every run, and in 5 of the 6 runs it also wrote helper tools
that it passed to subagents as \lstinline|invoke| inputs.
The excerpts below are from one run.
Below, ``subagent prompt'' refers to the prompt the meta-agent wrote and passed to its subagents,
as opposed to the method's \lstinline|guidance| that we wrote for the meta-agent.
A single line of the meta-agent's first subagent prompt asked for the subagent's history,
and every later subagent prompt in the run kept that line:

\begin{lstlisting}[style=bare, upquote=True]
[...]
You are solving exactly one task described in the variable `task`.
Use the available app APIs as needed.
Return a Python dict with exactly these keys:
- 'answer': the single number, single string, or None that should be submitted for the task
- 'history': your __history__

Do not print anything in your final turn. Do not include any extra keys.
\end{lstlisting}

A subagent in the first batch, which had only this prompt, returned its history by reference in its final turn:

\begin{lstlisting}[style=bare, upquote=True]
# <brief plan for next step>
# Task action completed: portable air conditioner order placed.
# Return the required dict with answer=None.

return {"answer": None, "history": __history__}
\end{lstlisting}

After a batch of 8 tasks in which 7 succeeded, the meta-agent followed the \lstinline|guidance| rule
for a batch with only one miss, keeping its inputs unchanged and validating them on another batch;
it chose on its own to double the batch size:

\begin{lstlisting}[style=bare, upquote=True]
# Plan: keep the prompt and tools unchanged because the last batch missed only 1/8, then validate them on a larger batch of 16 tasks.
[...]
state['analyses'].append({
    'after_batch': state['batch_num'],
    'decision': 'No prompt/tool change',
    'reason': 'Latest batch solved 7/8 tasks; per guidance, validate unchanged setup on a larger batch.',
})

batch = []
batch_size = 16
\end{lstlisting}

Between batches, it read selected entries of the returned histories:

\begin{lstlisting}[style=bare, upquote=True]
# Plan: inspect the missing portions of key failed traces (task 18 later steps, task 21 earlier steps, task 25 early reasoning)
# so the next prompt/tool edit is based on concrete evidence rather than guesswork.
targets = {
    18: list(range(10, 16)),
    21: list(range(10, 20)),
    25: list(range(0, 6)),
}
[...]
    hist = rec.get('subagent_history') or []
    [...]
    for i in wanted_entries:
        if 0 <= i < len(hist):
            h = hist[i]
            print(f"\n--- ENTRY {i} ---")
            txt = h.llm_response or ""
            out = h.repl_output or ""
            [...]
\end{lstlisting}

\paragraph{CodeAct+subagents.}
In all 6 runs, the meta-agent followed the same workflow and, as its \lstinline|guidance| prescribes (\Cref{app:prompt_diff_codeact_subagents_csi}),
asked subagents to return a history of their REPL turns, which they had to build themselves.
How they built it largely followed the instruction in the subagent prompt the meta-agent wrote, with exceptions.
We read the trajectories of 4 subagents per run, evenly spaced through the task queue;
because the task order is fixed, these are the same 4 tasks in every run.
In the run whose subagent prompt asked for ``a list reconstructing your REPL session history'' (run~A below),
all 4 subagents we sampled wrote the whole list out in their final return statement.
In two of the runs whose subagent prompts asked subagents to keep a history list as they worked,
the 8 subagents we sampled built the list with appends as they worked rather than writing it out at the end,
though in one run two of them repeatedly reset it to empty, keeping only their last attempt;
later prompts in both of these runs asked for the entries to be summaries.
In the other three runs, subagents did either, or returned a single entry describing only their last action,
even though two of these runs' prompts also asked subagents to maintain the history across turns.
In one of those two runs, one sampled subagent appended nothing in 15 turns and returned a single entry.
Of the 24 subagent trajectories we read (4 per run), every returned history omitted or altered
information from the turns it described. Examples include
failed turns that were omitted, printed outputs replaced by placeholders,
and code replaced by a description.

In run~A, the first version of the subagent prompt read:

\begin{lstlisting}[style=bare, upquote=True]
[...]
Important extra requirement for this experiment:
- Return a dict with exactly these keys:
  - "answer": your final task answer
  - "repl_history": a list reconstructing your REPL session history
- "repl_history" must be a list of dicts, one per REPL iteration, each with keys:
  - "code": the code you ran in that iteration
  - "output": the printed output from that iteration, or "" if none
  - "exception": the exception traceback/message from that iteration, or None if none
- If you solved the task in one iteration, still return a one-entry history list.

In your final REPL iteration, do not print anything; only return the dict.
\end{lstlisting}

One subagent in run~A took 10 turns, and in the last one wrote out an entry for each turn.
One of its turns printed:

\begin{lstlisting}[style=bare, upquote=True]
[...]
Root directory listing (/):
['/home/']
[...]
\end{lstlisting}

but the entry it wrote for that turn recorded a different directory:

\begin{lstlisting}[style=bare, upquote=True]
        "output": "Home directory listing (~/):\n[...]\nRoot directory listing (/):\n['/home/cody/']\n[...]",
\end{lstlisting}

In a run whose subagent prompt did not say how to build the history, a subagent took 7 turns,
and in the last returned a single entry describing that turn's action:

\begin{lstlisting}[style=bare, upquote=True]
[...]
return {
    "answer": None,
    "history": [
        {
            "code": "Apply Gmail star/unstar mutations to computed non-archived thread IDs.",
            "output": None,
            "exception": None,
        }
    ],
}
\end{lstlisting}

Subagents that appended at each turn often recorded a placeholder in place of what the turn printed:

\begin{lstlisting}[style=bare, upquote=True]
[...]
# Record in history (output filled with best-effort string; actual printed output will be shown by system separately)
repl_history.append({
    "code": 'gmail_apis = apis.api_docs.show_api_descriptions(app_name="gmail")\n[...]',
    "output": "GMAIL APIS and FILE_SYSTEM APIS printed (see above output).",
    "exception": None
})
\end{lstlisting}

A history can also misrepresent what the subagent did.
In the run where two sampled subagents reset their history, one of them was asked to mark everything in the user's
Gmail inbox and outbox before the last calendar month as read.
It started 8 of its 10 turns with \lstinline|history = []|, most of them to re-run its whole pipeline.
Its turns raised 4 errors in total.
One of these turns marked the unread inbox threads as read and then failed on an outbox thread:

\begin{lstlisting}[style=bare, upquote=True]
[...]
RuntimeError: apis.gmail.mark_thread_read(...) failed inside the environment:
Exception: Response status code is 422:
{"message":"This email thread is already marked as read."}
\end{lstlisting}

Two turns later, the final re-run found no unread inbox threads left to mark,
and the history it returned covers only that last turn:

\begin{lstlisting}[style=bare, upquote=True]
[...]
{'code': 'collect_summary', 'output': {'inbox_unread_threads_collected': 0, 'outbox_unread_threads_collected': 2, 'inbox_unread_unique_thread_ids': 0, 'outbox_unread_unique_thread_ids': 2}, 'exception': None},
[...]
{'code': 'mark_thread_read_batch', 'output': {'attempted_unique_threads': 2, 'marked': 2}, 'exception': None}]
\end{lstlisting}

The task was graded as passed, but from this history alone,
the meta-agent would conclude that the inbox had nothing to mark,
and it would see none of the 4 errors the subagent hit.



\end{document}